\documentclass[a4paper, 11pt]{article}

\usepackage{graphicx}
\usepackage{amsmath}
\usepackage{amsfonts}
\usepackage{amssymb}
\usepackage{authblk}
\usepackage{graphicx}
\usepackage{subfigure}
\usepackage{amsmath,amssymb,listings}
\usepackage{color}
\usepackage{url}
\usepackage{bm}
\usepackage{dsfont}
\usepackage[normalem]{ulem}
\usepackage{epsfig,amsmath,amssymb,mathrsfs}
\usepackage[margin=1in]{geometry}
\usepackage{graphicx,color}
\usepackage{setspace,lineno,epsfig}
\usepackage{algorithm}

\newtheorem{pro}{Proposition}

\newtheorem{Thm}{Theorem}

\usepackage{tikz}
\usetikzlibrary{arrows,backgrounds,snakes,patterns}
\usetikzlibrary{shapes,arrows,chains}
\usepackage{verbatim}
\usepackage{booktabs}

\begin{document}

\title{A Variational Image Segmentation Model based on Normalized Cut with Adaptive Similarity and Spatial Regularization
\thanks{This work was
supported by The National Key Research and Development Program of China (2017YFA0604903). Liu was also supported by the National Natural Science Foundation of China (No. 11871035).}}

\author{Faqiang Wang~$^{\dag}$, Cuicui Zhao~$^{\dag}$, Jun Liu~$^{\dag}$,  Haiyang Huang \thanks{Laboratory of Mathematics and Complex Systems (Ministry of Education of China), School of Mathematical Sciences, Beijing Normal University, Beijing, 100875, People's Republic of China.
(Jun Liu: jliu@bnu.edu.cn). }}
\date{}
\maketitle

\begin{abstract}
Image segmentation is a fundamental research topic in image processing and computer vision.
In the last decades, researchers developed a large number of segmentation algorithms for various applications. Amongst these algorithms, the Normalized cut (Ncut) segmentation method is widely applied due to its good performance.
The Ncut segmentation model is an optimization problem whose energy is defined on a specifically designed graph.
Thus, the segmentation results of the existing Ncut method are largely dependent on a pre-constructed similarity measure on the graph since this measure is usually given empirically by users. This flaw will lead to some undesirable segmentation results. In this paper, we propose a Ncut-based segmentation algorithm by integrating an adaptive similarity measure and spatial regularization.
The proposed model combines the Parzen-Rosenblatt window method, non-local weights entropy, Ncut energy, and regularizer of phase field in a variational framework. Our method can adaptively update the similarity measure function by estimating some parameters. This adaptive procedure enables the proposed algorithm finding a better similarity measure for classification than the Ncut method.
We provide some mathematical interpretation of the proposed adaptive similarity from multi-viewpoints such as statistics and convex optimization.
In addition, the regularizer of phase field can guarantee that the proposed algorithm has a robust performance in the presence of noise, and it can also rectify the similarity measure with a spatial priori.
The well-posed theory such as  the existence of the minimizer for the proposed model is given in the paper.
Compared with some existing segmentation methods such as the traditional Ncut-based model and the classical Chan-Vese model, the numerical experiments show that our method can provide promising segmentation results.
\end{abstract}



\pagestyle{myheadings}
\thispagestyle{plain}


\section{Introduction}\label{intro}

Image segmentation is a low level computer vision task which is to partition an image into several regions according to different requirements in applications. It has been studied and investigated in many works \cite{all2}, and many methods have been proposed during the last decades. Roughly speaking, there are two popular techniques in image segmentation filed in recent years: the handcraft based model and the learning based method.

The learning based method is very successful and popular for image segmentation recently, especially for deep learning neural network based works \cite{nn1,nn2,nn3}. If there are plenty of training samples, these models can produce some impressive results.
 However, to apply these techniques, one always needs a large amount of labeled data to train some desirable networks, but this is unavailable in some real applications. In this paper, we focus on the other method.

The handcraft based technique involves designing a model with some artificial prioris, such as the pixels in the same region should share the same mean, variance, probability density and so on. Some representative handcraft based models include snakes and active contour models \cite{snakes9,snakes3}, which is to evolve a curve to detect the edges of images by developing an evolution equation. It is well-known that this method is easy to get stuck in a local minimization due to the proposition of evolution equation. Thus, the results of
segmentation are largely dependent on the given initial curves. Expectation Maximization (EM) \cite{9aj} based Markov Random Field \cite{mrf} model \cite{emref,8jlhz} is another type of handcraft algorithm. In these models, pixels in an image can be regarded as some samples taken from a random variable whose probability distribution can be captured by a parametric mixture model. Then the segmentation could be transferred to  a statistical parameter estimation problem. However, the parametric model framework may limit the applications of mixture models, since the real data  may not follow a standard distribution. In some sense, these  two handcraft methods can be unified in a variational framework.


The variational approach is a popular and successful technique for image segmentation. In this method (e.g. \cite{mumford,varseg,cvmt,phase,emref,view21}),
segmentation results usually can be obtained by minimizing a cost functional which consists of data term (segmentation similarity) and regularization term (smoothness).
The regularization term is particularly useful to enhance the smoothness of the results when the images contain noise.
For example, the Total Variation (TV) regularization penalizes the length of region contours \cite{11,mumford,1cv,cv2} to get some clear segmentation boundaries, and the H$^{1}$ regularizer \cite{varseg,liuzheng} (Dirichlet energy of the phase field models \cite{phase,phase2})
can enhance the smoothness of the classifications.
This way, the algorithm is more robust to noise.
 Mathematically, the regularization technique can reduce the solution spaces such that the ill-posed segmentation problem becomes well-posed. Meanwhile, the regularizer can also improve the accuracy of segmentation and enhance the robustness of the results under noise.
The data term appeared in the cost functional commonly formulates the clustering energy of image, such as k-means \cite{kmeans,1cv} and spectral clustering methods (e.g. normalized cut) \cite{3sm,15wr,speseg1}. The famous Chase-Vese (CV) \cite{1cv,cvmt} model takes k-means clustering energy as data term, which has a good segmentation performance for some center-separable data. Let $I:\Omega \rightarrow \mathbb{R}$ be an image defined on an open bounded set $\Omega\subset \mathbb{R}^{2}$. The CV model \cite{1cv} is a piecewise constant approximation of two-phase Mumford-Shah segmentation model \cite{mumford}, and the energy can be written as
\begin{equation*}\label{cv1}
\mathcal{F}(\phi,\mu_{1},\mu_{2})
=\lambda_{1}\int_{\Omega}|I(x)-\mu_{1}|^{2}H(\phi(x))dx+\lambda_{2}\int_{\Omega}|I(x)-\mu_{2}|^{2}(1-H(\phi(x)))dx+\int_{\Omega}|\nabla H(\phi(x))|dx,\\
\end{equation*}
where $\phi$ is the signed distance function and $H$ is the Heaviside function. $\mu_{1},\mu_{2}$ are two unknown means of the pixels in regions, and $\lambda_{1}>0, \lambda_{2}>0$ are two given control parameters.

The traditional method to solve CV model is to evolve the level-set function by gradient flow to obtain the segmentation results.
In recent years, many fast algorithms \cite{7,13ca,sbm,29} have been developed for this problem. There are many advantages of the CV model such as flexibility, stability, and robustness. For example, the spatial priori is very easy to be plugged into the segmentation algorithm in a variational framework. However, the CV model cannot address the clustering problem of non-central distributed data, such as the nested double-moon dataset since it is a center-based clustering algorithm.

To classify some non-central distributed data, the spectral clustering \cite{2uvl} is proposed. The key idea of spectral clustering is to transform the data points into a feature space, in which the data can be easily classified by some algorithms such as center-based K-means clustering. Spectral clustering is derived from spectral graph theory \cite{4cf,graphtheory2}; it can be reduced to a min-cut problem on a specifically designed graph. 
Moreover, spectral clustering can be regarded as a manifold regularization \cite{view22} as well, since this term can catch geometric features, which is widely used in the machine learning field \cite{manifold3,manifold1,manifold2}. Due to its powerful nonlinear separable ability, spectral clustering is widely used in segmentation methods \cite{speseg1,speseg2,3sm,dncut2}. In these models, {an image is represented as an undirected weighted graph $G=<\mathbb{V},\mathbb{E},w>$, where $\mathbb{V}$ is a set of pixels, $\mathbb{E}$ is an edges set, and $w$ is a real-valued scalar similarity measure function (SMF) on the edges. Usually, $w$ is given and fixed in advance.} As graph $G$ is undirected, the SMF $w$ is symmetric. {Then the image segmentation can be casted as a min-cut problem of a specific graph}. Luxburg \cite{2uvl} gave a comprehensive review of spectral clusterings, and many technical details, such as the construction of similarity graph and some tricks of parameters choices, could be found in it. Wu and Leahy \cite{15wr} established a cut-based segmentation method. Their model  performs well. However, their algorithm favors grouping pixels into small sets due to the ratio bias of cuts, which is undesirable in some applications.
To resolve this problem, Hagen and Kahng presented a Cheeger cut criterion based clustering method \cite{ratiocut22,12hk}, which balanced the ratios of cuts and showed a better performance. Another solution of the ratio bias is the well-known Normalized cut (Ncut) method \cite{3sm}, which overcomes this drawback by introducing a normalized cost functional
\begin{equation*}\label{N-Cut}
Ncut(\mathbb{A},\mathbb{B})=\frac{cut(\mathbb{A},\mathbb{B})}{assoc(\mathbb{A},\mathbb{V})}+\frac{cut(\mathbb{A},\mathbb{B})}{assoc(\mathbb{B},\mathbb{V})},
\end{equation*}
where $(\mathbb{A},\mathbb{B})$ is a partition of graph $G$, and $assoc(\mathbb{A},\mathbb{V})=\underset{x\in\mathbb{A}}{\sum}
~\underset{y\in\mathbb{V}}{\sum}w(x,y)$ is a normalization factor.

The normalization factor $assoc(\cdot,\cdot)$ can improve the performance of image segmentation since
 small classes can be partly avoided in this model. However, this binary optimization of the Ncut is NP-Hard. Fortunately, this problem can be transferred to a generalized-eigenvalue problem by relaxation \cite{3sm} as
\begin{equation}\label{Ncut_eigen}
(\bm {D}-\bm {W})\bm{f}=\lambda \bm {D}\bm{f},~s.t.~\bm{f}^{'}\bm {D}\textbf{1}=0.
\end{equation}
{Here the relaxation function $\bm{f}$ of a binary variable can be used to label the segmentation, $\bm{W}$ is a similarity matrix of the graph $G$, and $\bm {D}$ is a degree matrix \cite{2uvl} which is a diagonal matrix with $\bm{D}_{ii}=\sum_{j}\bm{W}_{ij}$.}

Together with normalizing condition, the relaxation problem \cite{bn} can be rewritten as
\begin{equation}\label{ncut1}
\min_{\scriptstyle \bm{f}^{'}\bm {D}\bm{f}=1 \atop \scriptstyle \bm{f}^{'}\bm {D}\textbf{1}=0}\bm{f}^{'}(\bm {D}-\bm {W})\bm{f}.
\end{equation}

Then, it is easy to get the continuous version of (\ref{ncut1})
\begin{equation}\label{ncut2}
\min_{f\in \mathbb{F}}\left\{\int_{\Omega}\int_{\Omega}w(x,y)(f(x)-f(y))^{2}dxdy\right\}\footnote{If $\Omega$ is a discrete set, then $\int=\sum$.},
\end{equation}
where $\displaystyle\mathbb{F}=\displaystyle\{f:\Omega\rightarrow \mathbb{R}~|~\displaystyle\int_{\Omega}f(x)d(x)dx=0, \displaystyle\int_{\Omega}f^{2}(x)d(x)dx=1\displaystyle\}$ {and $d(x)=\displaystyle\int_{\Omega}w(x,y)dy.$}

{In fact, equation (\ref{ncut2}) is a nonlocal Dirichlet energy \cite{phase,phase2} with constraints on a graph.} Szlam and Bresson \cite{5asx}, Buhler and Hein \cite{6tbmh} showed the relationship between spectral clustering and nonlocal total variation \cite{11gs}, respectively. They provided some theoretical results of spectral clustering as well.

Since the Ncut-based problem can be relaxed to an eigenvalue system, it can be solved efficiently. Therefore, many Ncut-based segmentation models have been proposed. Yu and Shi \cite{yushi} established a Ncut-based model by giving some labels as a priori. Mathematically, it is the Ncut model with a linear homogeneous equality constraint. In addition, it was extended to a model with non-homogenous equalities constraints by Eriksson et al. \cite{erik}. Bernard et al. \cite{dncut2} proposed a new framework to solve the Ncut problem with priori and convex constraint by the Dinkelbach method \cite{dncut1}. 
Although these spectral clustering based methods have been proven to perform well, they are still sensitive to noise, since they lack of spatial priori information and regularization \cite{1cv,mumford,varseg}. Moreover, some additional post processes need to be taken to get good results. Tang et al. \cite{ncmrf} combined MRF regularization with Ncut to show better robustness. However, the KNN affinity (similarity) construction was adopted in this model with pre-specified window parameter $K$, this may sometimes lead to bad segmentation results if one chose improper parameters. On the spatial regularization problem, Yu et al. \cite{pfe} employed an $L^{1}$-regularized energy term in cut-based formulation to promote sparse solutions, and the affinities similarity of Ncut process \cite{3sm} was adopted in a piecewise-flat embedding model, which should be given in advance.

In fact, most spectral clustering based models apply KNN-based similarity graph construction with a given parametric similarity measure function (SMF). For different scales of data,  choosing a proper parameter in SMF is intractable. There are less works and theoretical results in this respect.

In this work, we establish a Ncut-based variational segmentation model which can adaptively update the SMF. The proposed model combines Parzen-Rosenblatt window method \footnote{Parzens-Rosenblatt window method is also termed as Kernel Density Estimation (KDE).} \cite{parzen1,parzen2}, non-local weights entropy, Ncut energy and regularizer of phase field in variational framework. The model can update the bandwidth of Parzen-Rosenblatt window during the iteration and enable our model to have an adaptive SMF. Moreover, the regularization of phase will enhance spatial smoothness of the spectrum vector. This spatial priori information  can also improve the SMF.

To obtain an adaptive similarity, we consider the image intensities as some realizations of a random vector, and adopt the Parzen-Rosenblatt window method to estimate its probability density function, which is a Gaussian Mixture Model (GMM). Inspired by GMM \cite{10gmm} and the EM \cite{9aj} method, we obtain a functional related to pixels similarities. To combine these totally different methods, we propose a general Ncut-based segmentation algorithm in a variational framework. To enhance the robustness of the proposed models, we adopt the H$^{1}$ (Dirichlet energy, see e.g. \cite{liuzheng,phase}) and total variation (TV) as spatial regularization.
Though the H$^{1}$ regularization shows higher computation efficiency, it smoothes the boundary of an object as well. TV regularizer can penalize the length of contour \cite{11} and get sharp segmentation boundaries \cite{1cv,cv2,phase2}. In fact, Ncut is a linear system based method under relaxation. However, the introduced TV regularizer is nonlinear and it makes the model no longer a linear eigenvalue problem. Here, we introduce Dinkelbach algorithm \cite{dncut1,dncut2} to solve our proposed models.

{The main contribution of this paper includes:
\begin{enumerate}
\item We
construct an adaptive similarity measure which combines the optimal Parzen-Rosenblatt window and spatial priori knowledge (regularization of phase field) for Ncut-based segmentation. In our method, the similarity matrix can be updated adaptively, and thus we can get a good segmentation measure. The results of image segmentation are better and more stable than Nut's.
\item We propose a segmentation model by combining the kernel density estimation, EM algorithm, spectral graph energy (nonlocal Dirichlet energy with constraints) and TV regularizer, which integrates many superiorities of different methods. Compared with some existing Ncut-based methods and classical Chan-Vese model, the numerical experiments show that our algorithm can achieve desirable segmentation performance.
\item The existence of the minimizer for the proposed model is mathematically shown.
\end{enumerate}
}

The rest of this paper is organized as follows. 
Section 2 describes the proposed variational Ncut-based segmentation models with adaptive similarity and regularization, 
 and gives the existence of the minimizer of the  proposed model as well. Section 3 shows the algorithms and the details in implementation. In section 4, we design some numerical experiments to demonstrate our models. We summarize our methods and conclude in section 5.

\section{The Proposed Method}
\label{sec_3}
\subsection{Statistical Methods}
\label{sec_3.1}

In this section, we shall propose two Ncut-based segmentation models with adaptive similarity and spatial regularization in a variational framework. To obtain a similarity measure, we approximate the image distribution by the Parzen-Rosenblatt window method \cite{parzen1,parzen2} in which the kernel functions are Gaussian types. That is, any distribution can be very well approximated by GMM. 
Inspired by parameters estimation of GMM and the EM algorithm, we introduce an auxiliary variable which can measure the similarity of pixels. Then we can formulate the process of parameters estimation as an alternating optimization. In our model, we combine Parzen-Rosenblatt window method, Ncut energy and regularizer of phase field, and thus the similarity function can be determined by the cost functional itself. In the next, we will present our method.

\subsubsection{Image Likelihood by Parzen-Rosenblatt Window Method}
\label{sec_3.2}
{In this part, we estimate image distribution by Parzen-Rosenblatt window method \cite{parzen1,parzen2}, in which the kernel function is Gaussian. By using the idea of the EM algorithm, we will get a functional which measures the similarity of pixels.}

Let $\Omega$ be a discrete set and $|\Omega|=N$. We assume that the intensity values $\bm o=(I(x_1),\cdots,I(x_N))$ is a realization of a random vector $\mathscr{O}=(\mathscr{O}_1,\cdots,\mathscr{O}_{N})$, whose components are independent identically distributed (i.i.d). Then the probability density of $\mathscr{O}$ can be approximately expressed as
\begin{equation*}
p(z)=\frac{1}{|\Omega|}\int_{\Omega}K_{h}(z-I(y))dy,
\end{equation*}
where $K_{h}$ is chosen as
\begin{equation*}
K_{h}(x)=\frac{1}{\sqrt{2\pi}h}e^{-\frac{x^{2}}{2h^{2}}}.
\end{equation*}
When $h\rightarrow 0$, then $K_{h}(x)\rightarrow \delta(x)$ in the sense of distributions, and this approximation is indeed the normalized histogram of $I$ or empirical distribution of image.

By substituting $K_h$ into $p(z)$, we get
\begin{equation}\label{hist}
p(z)=\frac{1}{|\Omega|}\int_{\Omega}\frac{1}{\sqrt{2\pi}h}e^{-\frac{(z-I(y))^{2}}{2h^{2}}}dy.
\end{equation}

Obviously, $p(z)$ is a Gaussian mixture distribution parameterized by $h$, so we deal with it by the EM process as the following. 

\subsubsection{Adaptive Similarity Functional}
\label{sec_3.3}
By the i.i.d. assumption, we get the related negative log-likelihood function
\begin{equation}\label{hist_para}
L(h)=-\int_{\Omega} \ln (\frac{1}{|\Omega|}\int_{\Omega}\frac{1}{\sqrt{2\pi}h}e^{-\frac{(I(x)-I(y))^{2}}{2h^{2}}}dy)dx.\\
\end{equation}

{The parameter $h$ can be estimated by minimizing the above negative log-likelihood function (\ref{hist_para}),
this is equivalent to a Maximum Likelihood Estimation (MLE).
For an efficient computation, we adopt the EM algorithm. Let us introduce a hidden random vector $\mathscr{U}=(\mathscr{U}_1,\cdots,\mathscr{U}_N)$, whose realization $\bm u=(u_1,\cdots,u_N)$ indicates that the sample $\bm o$ comes from the $\bm u$-th component of the Gaussian mixture. Then we have the complete data as $(\mathscr{O},\mathscr{U})$, in which a realization of $\mathscr{U}_i$, denoted as $u_i$, implies that the sample $o_i$ is produced by the $u_i$-{th} Gaussian distribution.} Then according to a standard EM process \cite{9aj,8jlhz}, we have
\begin{equation}\label{hist_em}
L(h)=Q(h;h^{t-1})-H(h;h^{t-1}),
\end{equation}
where $t$ is an iteration number and
\begin{equation*}
Q(h;h^{t-1})=-\int_{\Omega\times\Omega}\ln\left(\frac{1}{|\Omega|}p_y\left(I(x);h\right)\right)p(y|I(x);h^{t-1})dxdy,
\end{equation*}
and
\begin{equation*}
H(h;h^{t-1})=-\int_{\Omega\times\Omega} \ln\left( p(y|I(x);h)\right)p(y|I(x);h^{t-1})dxdy.
\end{equation*}
It is not difficult to check that $L(h)\leqslant Q(h;h^{t-1}), \forall h^{t-1}$, by using the fact that $H(h;h^{t-1})\geqslant 0$ (Jensen's inequality).
Details of derivation of (\ref{hist_em}) can be found in references \cite{9aj,8jlhz} \emph{etc.}.


Using the fact that
\begin{equation*}
p_y\left(I(x);h\right)=\frac{1}{\sqrt{2\pi}h}e^{-\frac{(I(x)-I(y))^{2}}{2h^{2}}},
\end{equation*}
and we plug it into $Q$, and then
\begin{equation*}
Q(h;h^{t-1})=-\small{\int_{\Omega\times\Omega} \ln(\frac{e^{-{(I(x)-I(y))^{2}}/{(2h^{2})}}}{\sqrt{2\pi}h|\Omega|})p(y|I(x);h^{t-1})dxdy},
\end{equation*}
where $\int_{\Omega}p(y|I(x);h^{t-1})dy=1$.

{
Notice that $p(y|I(x);h^{t-1})$ in $Q(h;h^{t-1})$ represents the probability that the pixel $I(x)$ belongs to $y$-th group. Here, since the number of groups is equal to the number of image pixels, this probability can be used to measure the similarity between $I(x)$ and $I(y)$. For these reasons, we introduce an auxiliary function $w:\Omega\times\Omega\rightarrow\mathbb{R}, w(x,y)=p(y|I(x);h^{t-1})$ containing parameter $h$, then the parameters estimation (\ref{hist_em}) can be converted to a minimization process by using the idea of EM as follows}
\begin{equation}\label{optimal_EM}
\min\limits_{h\in\mathbb{H},w\in \mathbb{C}_{1}}\mathcal{E}_{1}(h,w),
\end{equation}
where
\begin{equation*}
\mathcal{E}_{1}(h,w)=\displaystyle\int_{\Omega\times\Omega}(\frac{(I(x)-I(y))^{2}}{2h^{2}})w(x,y)
+\ln(\sqrt{2\pi}h|\Omega|)w(x,y)+ w(x,y)\ln w(x,y)dxdy,
\end{equation*}
 and $\mathbb{C}_{1}=\{w:\Omega\times\Omega\rightarrow\mathbb{R}|0\leq w(x,y)\leq  1,$ $\int_{\Omega}w(x,y)dy$ $=1, \forall x\in\Omega\},$ $\mathbb{H}=\{h|0<h_{min}\leq h\leq h_{max}<+\infty\}.$

{In fact, $\mathcal{E}_{1}$ is the exact formulation of $Q-H$ by replacing  $w(x,y)=p(y|I(x);h^{t-1}).$ Also, the lower bound of $\mathcal{E}_{1}(h,w)$ with respect to $w$ is equal to $L(h)$ (here $L(h)$ is the negative likelihood function defined in (\ref{hist_para})).
Thus $\mathcal{E}_{1}(h,w)$ can be seen as an upper bound function of $L(h)$. We will see the exact relationship in the next section.
Moreover, the functional of two variables $\mathcal{E}_{1}(h,w)$ is easier to be optimized. In addition, one can verify that (\ref{optimal_EM}) has the same estimator as the original parameters estimation problem (\ref{hist_para})\cite{8jlhz}.}

The minimization problem (\ref{optimal_EM}) can be efficiently solved by the alternating scheme
\begin{equation}\label{add_liu}
\left\{
\begin{array}{rl}
h^{t}=&\underset{h\in\mathbb{H}}{\arg\min} ~~\mathcal{E}_{1}(h,w^{t-1}),\\
w^{t}=&\underset{w\in \mathbb{C}_{1}}{\arg\min} ~~\mathcal{E}_{1}(h^{t},w).\\
\end{array}
\right.
\end{equation}

{As for the EM iteration scheme (\ref{add_liu}), it has been proven the energy corresponding to this problem is decreasing with respect to the parameter $h$ \cite{8jlhz}.} By using the Lagrangian multiplier method, we can easily get a closed-form projection of $w(x,y)$ on probabilistic simplex $\mathbb{C}_1$ in the second subproblem of (\ref{add_liu})
\begin{equation}\label{simi}
w^t(x,y)=\frac{1}{S(x)}e^{-\frac{(I(x)-I(y))^{2}}{2(h^t)^{2}}},
\end{equation}
where $\displaystyle S(x)=\int_{\Omega}e^{-\frac{(I(x)-I(y))^{2}}{2(h^t)^{2}}}dy$ serves as a normalization factor.

{In summary, by Parzen-Rosenblatt window method, we have an approximation of image intensity histogram, which is a GMM.
Inspired by the GMM and EM parameters estimation, we obtain a functional with two variables. One variable is the model parameter $h$ to be estimated; the other variable $w$ (\ref{simi}) can be used to measure the similarity between image pixels, which has the similar form of the commonly used Gaussian similarity function \cite{2uvl}.
Besides, the similarity (\ref{simi}) is obtained by the variational model (\ref{optimal_EM}) itself, which can easily be casted in a variational framework with regularization.}

Since the existence of normalization factor $S(x)$ in the similarity function obtained by (\ref{simi}), the SMF $w$ is asymmetric. However, in the Ncut method, the symmetrization of similarity matrices would significantly simplify the algorithm. To incorporate the symmetrization into the Ncut structure,
we would like the similarity function $w(x,y)$ to be symmetric. To this end, we project the asymmetric SMF to a convex set $\mathbb{C}_{2}$
which is formed by symmetric functions. For this projection, it is not difficult to get the following result:

{\begin{pro}\label{pro_1}
Let $\mathbb{C}_{2}=\{s:\Omega\times\Omega\rightarrow \mathbb{R}^{+},s(x,y)=s(y,x),\forall x\in\Omega,y\in\Omega\}$,
then $\mathbb{C}_{2}$ is convex, and the projection of a scalar function $w:\Omega\times\Omega\rightarrow \mathbb{R}^{+}$ 
onto $\mathbb{C}_{2}$ is $s(x,y)=\frac{w(x,y)+w(y,x)}{2}$.
\end{pro}}

It becomes difficult to address the problem when it is loaded with both the earlier mentioned probabilistic simplex constraint and the symmetry condition (i.e. $w\in\mathbb{C}_1\cap\mathbb{C}_{2}$ ).
In fact, this problem does not have a closed-form solution.
However, these two constraints are both convex so that we can use the projection method successively \cite{pj1,pj2,projcon}. Thus, in real implementations, we just need to run equation (\ref{simi}) and proposition \ref{pro_1} successively.



\subsection{Convex Optimization Interpretation for $w$}
{In fact, $w$ appeared in EM algorithm is a Fenchel's dual variable from the view point of convex optimization, and (\ref{optimal_EM}) can be deduced from the following proposition:
\begin{pro}\label{pro_dual}
The functional
$$
\mathcal{J}(u)=\int_{\Omega}\ln(\int_{\Omega}e^{u(x,y)}dy)dx,
$$
is convex with respect to $u$, and
$$\mathcal{J}^{**}(u)=\mathcal{J}(u)=\max\limits_{w\in\mathbb{C}_{1}}\left\{\int_{\Omega\times\Omega}u(x,y)w(x,y)dxdy-\int_{\Omega\times\Omega}w(x,y)\ln w(x,y)dxdy\right\},$$
where $\mathcal{J}^{**}=(\mathcal{J}^{*})^{*}$ and $\mathcal{J}^{*}(w)=\max\limits_{u}\{<u,w>-\mathcal{J}(u)\}$ is the Fenchel-Legendre transformation of $\mathcal{J}$.
\end{pro}}

The proof of the Proposition is deferred to Appendix \ref{appendix_B0}.

Now, we use proposition \ref{pro_dual} to get equation \eqref{optimal_EM} from \eqref{hist_para}.

{
Let $u(x,y)=\frac{-(I(x)-I(y))^{2}}{2h^{2}}-\ln(\sqrt{2\pi}h|\Omega|)$. Then,
the MLE (\ref{hist_para}) becomes
\begin{equation}
\begin{array}{l}
\arg\min\limits_{h\in\mathbb{H}} L(h)\Longleftrightarrow
\arg\min\limits_{h\in\mathbb{H}} -\mathcal{J}(\frac{-(I(x)-I(y))^{2}}{2h^{2}}-\ln(\sqrt{2\pi}h|\Omega|))\\
\Longleftrightarrow
\arg\min\limits_{h\in\mathbb{H}} -\mathcal{J}^{**}(\frac{-(I(x)-I(y))^{2}}{2h^{2}}-\ln(\sqrt{2\pi}h|\Omega|))
\underset{\text{prop. \ref{pro_dual}}}{\Longleftrightarrow}\arg\min\limits_{h\in\mathbb{H},w\in \mathbb{C}_{1}}\mathcal{E}_1(h,w).\\
\end{array}
\end{equation}
}

{
The above derivation is based on the fact that $L(h)=-\mathcal{J}(u)=-\mathcal{J}^{**}(u)$, which gives the connection between EM step (\ref{optimal_EM}) and convex optimization. Obviously, $-\mathcal{J}^{**}$ is a special upper bound of $L$,
thus the minimizer of $L$ can be obtained by minimizing its tightly upper bound function $-\mathcal{J}^{**}$, which is the key idea of EM.
 It can be seen that $w$ appeared in the EM process is actually a dual variable in convex optimization. Thus, the EM algorithm of statistics for GMM is just a dual algorithm. }

{From the above analysis, we have a  Ncut energy coupled with an adaptive SMF cost functional. To establish a variational image segmentation model, we can take these terms and combine the spatial regularization into a variational framework.
}

\subsection{Variational Framework}
\label{sec_3.5}
\subsubsection{Some Definitions and Notations}

{Let $w:\Omega\times\Omega\rightarrow \mathbb{R}^{+}$ be a nonnegative smooth similarity function. $f:\Omega\rightarrow\mathbb{R}$ is bounded almost everywhere. Given $k_{\epsilon}:B_{\epsilon}\rightarrow\mathbb{R}^{+}$ is a smooth weighting function with $\int_{B_{\epsilon}}k_{\epsilon}(z)dz=1$, where $B_{\epsilon}=\{z\in\mathbb{R}^{2}:||z||<\epsilon\}$. Symbol ``*" is the convolution operator, i.e. $(k_{\epsilon}*f)(x)=\int_{B_{\epsilon}}k_{\epsilon}(y)f(x-y)dy$.}

\subsubsection{Normalized Cut based Segmentation Model with Adaptive Similarity and Spatial Regularization}
Based on the analyses in the previous sections, we propose
the variational normalized cut based segmentation model with adaptive similarity and spatial regularization as

\begin{equation}\label{va_ncash1}
\renewcommand\arraystretch{2}{
\begin{array}{l}
\min\limits_{w\in \mathbb{C},f\in \mathbb{F},h\in \mathbb{H}}\left\{\displaystyle\biggl.\int_{\Omega\times\Omega}\left\{\frac{(I(x)-I(y))^{2}}{2h^{2}}
+\ln(\sqrt{2\pi}h|\Omega|)\right\}w(x,y)dxdy+\displaystyle\biggl.\int_{\Omega\times\Omega}w(x,y)\ln w(x,y)dxdy\right.\\
+\left.\lambda\displaystyle\biggl.\int_{\Omega\times\Omega}w(x,y)[(k_{\epsilon}*f)(x)-(k_{\epsilon}*f)(y)]^{2}dxdy+\eta\mathcal{R}(f)\right\},
\end{array}}
\end{equation}
where $$\mathbb{C}=\{w\in L^{\infty}~:~0\leq w(x,y)\leq 1,\int_{\Omega}w(x,y)dy=1,w(x,y)=w(y,x),a.e. x,y\in \Omega\},$$ and $$\mathbb{F}=\{f~:~ |f(x)|<C,a.e. x\in\Omega, \int_{\Omega}f(x)d(x)dx=0,\int_{\Omega}d(x)f^{2}(x)dx=1\},$$ $\mathbb{H}=\{h~:~0<h_{min}\leq h\leq h_{max}<+\infty\},$ and $\lambda, \eta$ are two positive parameters which control the balance of each term in the cost functional.


Here we choose two popular regularizers in the field of computer vision: $\text{H}^{1}$ regularizer or called Dirichlet energy \cite{liuzheng,phase} of phase field $f$, which is $\mathcal{R}(f)=\int_{\Omega}||\nabla f(x)||^{2}dx$, and TV regularizer \cite{1cv,13ca,5asx,11gs} which is $\mathcal{R}(f)=\int_{\Omega}||\nabla f(x)||dx$.

{
In fact, the first two terms in the proposed model serve as EM process of Parzen-Rosenblatt window method based image likelihood, which produces an adaptive similarity measure $w(x,y)$. More specifically, the first term is the non-local Dirichlet energy \cite{phase,phase2} of image intensity by ignoring parameter estimation term, and the second term is a negative entropy regularizer of non-local weight $w$, which forces the similarity measure $w$ to be smooth. In fact, the entropic regularization has been used in many works, such as image inpainting \cite{entropy}, image segmentation \cite{8jlhz} and restoration \cite{liuzheng}. The third term is the Ncut energy, which is different from the non-local Dirichlet energy since the existence of the normalization and orthogonal constraints. These terms serve as clustering process and the last term formulates spatial regularization to make our segmentation results to be smooth and robust to noise.}

For convenience, we call the above NCut-based segmentation model with Adaptive Similarity and $\text{H}^{1}$ regularizer as NCAS$\text{H}^{1}$. Similarly, we use NCASTV to stand for the model of \eqref{va_ncash1} in which $\mathcal{R}$ is TV regularizer.

\textbf{Remark: }{ In the proposed methods, we take some convolution operators to the phase field $f$, which is beneficial to prove the existence of minimizers theoretically in a proper functional space \cite{liuzheng}. Let us emphasize that we cannot obtain the existence of minimizer if we drop  these convolution operators. The interested readers can find the reasons in the proof of theorem \ref{thm_1}.
In fact, a convolution between a smooth kernel $k_{\epsilon}$ and $f$ can be seen as a spatial regularization, which will enhance the segmentation performance as well. In real implementation to reduce computational cost, we can let $\epsilon\rightarrow 0$, then $k_{\epsilon}$ would be a delta function and the convolution operators would disappear.}

\subsubsection{Existence of the Minimizer for the Proposed Models}

In this section, we will prove the existence of the minimizers for the proposed models. We will show this for NCASTV only since both of NCASTV and NCAS$\text{H}^{1}$ have similar results by choosing some proper function spaces (BV and $\text{H}^{1}$) with the similar analyses method.

 Let us consider the following energy functional for the NCASTV model
\begin{equation*}
\renewcommand\arraystretch{2}{
\begin{array}{l}
\mathcal{E}(f,w,h)=\displaystyle\biggl.\int_{\Omega\times\Omega}\left\{\frac{(I(x)-I(y))^{2}}{2h^{2}}+\ln(\sqrt{2\pi}h|\Omega|)\right\}w(x,y)dxdy+\displaystyle\biggl.\int_{\Omega\times\Omega}w(x,y)\ln w(x,y)dxdy\\
+\lambda\displaystyle\biggl.\int_{\Omega\times\Omega}w(x,y)[(k_{\epsilon}*f)(x)-(k_{\epsilon}*f)(y)]^{2}dxdy+\eta\displaystyle\biggl.\int_{\Omega}||\nabla f(x)||dx.
\end{array}}
\end{equation*}

 Here we will show the existence of the minimizer for the NCASTV model in the following space
 \begin{equation*}
 \renewcommand\arraystretch{2}{
\begin{array}{lll}
 \mathbb{X}:=&\{(f,w,h):f\in BV({\Omega}), w\in L^{\infty}(\Omega\times\Omega),
 0\leq w\leq 1,\displaystyle\biggl.\int_{\Omega}f(x)dx=0, \displaystyle\biggl.\int_{\Omega}f^{2}(x)dx=1,\\
 &\displaystyle\biggl.\int_{\Omega}w(x,y)dy=1,|f(x)|<C,
 w(x,y)=w(y,x), a.e. x\in\Omega, y\in\Omega,
 0<h_{min}\leq h\leq h_{max}<+\infty\}.
 \end{array}}
 \end{equation*}

\textbf{Remark:} Please note that there is no $d(x)$ for the constraints $\int_{\Omega}f(x)dx=0$ and $\int_{\Omega}f^{2}(x)dx=1$ in $\mathbb{X}$. This is because $d(x)=\int_{\Omega}w(x,y)dy=1$ according to $w\in\mathbb{C}$.

\begin{Thm}\label{thm_1}
 There exists at least one solution $(f^*,w^*,h^*)\in\mathbb{X}$ for NCASTV model, i.e.
\begin{equation}\label{pf_ncastv}
(f^*,w^*,h^*)=\arg\min\limits_{(f,w,h)\in \mathbb{X}}\mathcal{E}(f,w,h).
\end{equation}
\end{Thm}

\textbf{Proof}:  The proof is deferred to Appendix \ref{appendix_B}.

\section{Algorithms}
\label{sec_4}

In this section, we describe the algorithms corresponding to the NCAS$\text{H}^{1}$ model and the NCASTV model. Both of the models are solved by alternating minimization algorithm. The normalization constraint $\int_{\Omega}d(x)f^{2}(x)dx=1$
appeared in the Ncut energy is kept by the Lagrangian method. For the updating of Lagrangian multiplier, we will not apply the simple gradient descent/ascent scheme since its convergence depends on the choice of time step. Here, an idea of Dinkelbach method \cite{dncut1}
is applied to keep this normalization constraint.
As for the orthogonal constraint $\int_{\Omega}d(x)f(x)dx=0$, we adopt projection method \cite{pj1,pj2,projcon}.

It is easy to find out that the H$^{1}$ based model is still an eigenvalue problem which can be solved efficiently as Ncut, but the TV based model is more difficult. The main difficulty is that the linear property of the algorithm is destroyed by the non-quadratic TV.
Though some splitting methods could be applied to TV, e.g. \cite{7,29,projcon}, the related subproblem would not be the exact Rayleigh quotient formula and it can not be solved by the standard eigenvector algorithms. Fortunately, this problem can be solved by the Dinkelbach algorithm. To have a consistent scheme, in this paper, we adopt the Dinkelbach algorithm \cite{dncut1,dncut2} to solve both of the proposed models.

\subsection{Algorithm for NCAS$\text{H}^{1}$ Model}

\label{sec_4.1}
NCAS$\text{H}^{1}$ model (\ref{va_ncash1}) can be directly minimized by alternating minimization algorithm, one may have the subproblems as below
\begin{equation*}
\left\{
   \renewcommand\arraystretch{2}{
  \begin{array}{rll}
\min\limits_{w\in \mathbb{C}_{2}}\max\limits_{\beta}&\left\{\displaystyle\biggl.\int_{\Omega\times\Omega}\left\{\frac{(I(x)-I(y))^{2}}{2h^{2}}+\ln(\sqrt{2\pi}h|\Omega|)\right\}w(x,y)dxdy\right.
+\displaystyle\biggl.\int_{\Omega\times\Omega}w(x,y)\ln w(x,y)dxdy\\
&+\displaystyle\biggl.\int_{\Omega}\beta(x)(1-\displaystyle\biggl.\int_{\Omega}w(x,y)dy)dx
+\left.\lambda\displaystyle\biggl.\int_{\Omega\times\Omega} w(x,y)[(k_{\epsilon}*f)(x)-(k_{\epsilon}*f)(y)]^{2}dxdy\right\},\\
\min\limits_{h\in\mathbb{H}}& \left\{\displaystyle\biggl.\int_{\Omega\times\Omega}\left\{\frac{(I(x)-I(y))^{2}}{2h^{2}}+\ln(\sqrt{2\pi}h|\Omega|)\right\}w(x,y)dxdy\right\},\\
\min\limits_{f\in\mathbb{F}}&\left\{\lambda\displaystyle\biggl.\int_{\Omega\times\Omega}w(x,y)((k_{\epsilon}*f)(x)-(k_{\epsilon}*f)(y))^{2}dxdy
+\eta\displaystyle\biggl.\int_{\Omega} ||\nabla f(x)||^{2}dx\right\},\\
     \end{array}}
\right.
\end{equation*}
where $\beta$ is the Lagrangian multiplier, and $\lambda$ and $\eta$ are the parameters.

As for the subproblem of $w$, we solve the corresponding optimization problem and then project the optimum onto $\mathbb{C}_{2}.$

According to the optimal condition, we have that
\begin{equation*}
\renewcommand\arraystretch{2}{
\frac{(I(x)-I(y))^{2}}{2h^{2}}+1+\ln(\sqrt{2\pi}h|\Omega|)+\ln w(x,y)+\lambda[(k_{\epsilon}*f)(x)-(k_{\epsilon}*f)(y)]^{2}-\beta(x)=0.
}
\end{equation*}
Then,
\begin{equation}\label{w_1}
w(x,y)= A(x)e^{\frac{-(I(x)-I(y))^{2}}{h^{2}}-\lambda [(k_{\epsilon}*f)(x)-(k_{\epsilon}*f)(y)]^{2}},
\end{equation}
where $$A(x)=e^{-1-\ln(\sqrt{2\pi}h|\Omega|)+\beta(x)}.$$

Since $\displaystyle\biggl.\int_{\Omega}w(x,y)dy=1$, we get
\begin{equation}\label{w_2}
A(x)=\frac{1}{\displaystyle\biggl.\int_{\Omega}e^{\frac{-(I(x)-I(y))^{2}}{h^{2}}-\lambda [(k_{\epsilon}*f)(x)-(k_{\epsilon}*f)(y)]^{2}}dy}
\end{equation}
by integrating both sides of the former equation (\ref{w_1}).
Plugging (\ref{w_2}) into (\ref{w_1}), then
\begin{equation}
w(x,y)=\frac{e^{\frac{-(I(x)-I(y))^{2}}{h^{2}}-\lambda [(k_{\epsilon}*f)(x)-(k_{\epsilon}*f)(y)]^{2}}}{\displaystyle\biggl.\int_{\Omega}e^{\frac{-(I(x)-I(y))^{2}}{h^{2}}-\lambda [(k_{\epsilon}*f)(x)-(k_{\epsilon}*f)(y)]^{2}}dy}.
\end{equation}

{Note that the SMF obtained by our  model combines image intensity information and spatial information (information of phase field $f$). With the update of phase field $f$ and parameter $h$, the SMF can be adaptively updated by the model itself to better fit the data.}

Since $w\in \mathbb{C}_{2}$, according to Proposition \ref{pro_1}, the projection of SMF $w$ (denoted as $w$ as well) is
{\begin{equation*}
w(x,y):=\frac{w(x,y)+w(y,x)}{2}.
\end{equation*}}

According to the optimal condition of subproblem of $h$, $h$ can be optimized by
\begin{equation}\label{update_h}
h^{2}=Proj_{\mathbb{H}}(\frac{\displaystyle\biggl.\int_{\Omega\times\Omega}w(x,y)[I(x)-I(y)]^{2}dxdy}{|\Omega|}).
\end{equation}

{The model parameter $h$ is determined by the above formula (\ref{update_h}). With the update of the similarity $w$, the Parzen-Rosenblatt window method based approximation (\ref{hist}) will be close to the real image density function.}

{Recall that the $f$-subproblem is
\begin{equation}\label{h1_f}
\min\limits_{f\in\mathbb{F}}\left\{\lambda\displaystyle\biggl.\int_{\Omega\times\Omega}w(x,y)((k_{\epsilon}*f)(x)-(k_{\epsilon}*f)(y))^{2}dxdy
+\eta\displaystyle\biggl.\int_{\Omega} ||\nabla f(x)||^{2}dx\right\},
\end{equation}}

{and $$\mathbb{F}=\{f~:\int_{\Omega}f(x)d(x)dx=0,\int_{\Omega}f^{2}(x)d(x)dx=1\},$$}

{Denote $z(x)=\sqrt{d(x)}f(x)$, then the $f$-subproblem can be converted to an eigenvalue problem with respect to $z$:
\begin{equation*}
\min_{\scriptstyle \int_{\Omega}z^2(x) dx=1\atop \scriptstyle \int_{\Omega}\sqrt{d(x)}z(x) dx =0}\left\{\lambda\displaystyle\biggl.\int_{\Omega\times\Omega}w(x,y)((k_{\epsilon}*\frac{z}{\sqrt{d}})(x)-(k_{\epsilon}*\frac{z}{\sqrt{d}})(y))^{2}dxdy
+\eta\displaystyle\biggl.\int_{\Omega} ||\nabla \frac{z(x)}{\sqrt{d(x)}}||^{2}dx\right\}.
\end{equation*}
As for the constraint $\int_{\Omega}z^2(x) dx =1$, we can use the Lagrangian method to address it.}

{Define
$$
\mathbb{S}_{1}=\{z:\int_{\Omega}z^2(x) dx=1\},~~\mathbb{S}_{2}=\{z:\int_{\Omega}\sqrt{d(x)}z(x) dx =0\}.
$$
Then by the Lagrangian multiplier method, we have
\begin{equation}\label{dinkelbach_1}
\renewcommand\arraystretch{2}{
\begin{array}{l}
\min\limits_{z\in\mathbb{S}_{2}}\max\limits_{\mu}\left\{\lambda\displaystyle\biggl.\int_{\Omega\times\Omega}w(x,y)((k_{\epsilon}*\frac{z}{\sqrt{d}})(x)-(k_{\epsilon}*\frac{z}{\sqrt{d}})(y))^{2}dxdy
+\eta\displaystyle\biggl.\int_{\Omega} ||\nabla \frac{z(x)}{\sqrt{d(x)}}||^{2}dx+\mu\left(1-\int_{\Omega}z^2(x) dx\right)\right\}.
\end{array}}
\end{equation}}

{According to the first order optimal condition, we have the following linear equation:
$$
-\frac{\lambda}{\sqrt{d(x)}}\left(\hat{k}_{\epsilon}*\left(\triangle_{w}(k_{\epsilon}*\frac{z}{\sqrt{d}})\right)\right)(x)-\frac{\eta}{\sqrt{d(x)}}\triangle\frac{z(x)}{\sqrt{d(x)}}-\mu z(x)=0,
$$
where $\triangle_{w}$ is a graph Laplacian operator defined by
$$\triangle_{w}f(x)=-2\int_{\Omega} w(x,y)(f(x)-f(y))dy,$$
and $\hat{k}_{\epsilon}$ is the conjugate function of $k_{\epsilon}$, i.e $\hat{k}_{\epsilon}(x)=k_{\epsilon}(-x).$}

By multiplying $z(x)$ and integrating on both sides of the above optimization condition, we have
\begin{equation}
\mu=\displaystyle\biggl.\frac{-\lambda\displaystyle\biggl.\int_{\Omega}\frac{ {z}(x)}{\sqrt{d(x)}}\left(\hat{k}_{\epsilon}*\left(\triangle_{w}(k_{\epsilon}*\frac{{z}}{\sqrt{d}})\right)\right)(x)dx
-\eta\displaystyle\biggl.\int_{\Omega}\frac{{z}(x)}{\sqrt{d(x)}}\triangle\frac{{z}(x)}{\sqrt{d(x)}}dx}{\displaystyle\biggl.\int_{\Omega}z^{2}(x)dx}.
\end{equation}
By projection gradient method, and inspired by Dinkelbach method \cite{dncut1,dncut2}, we give the following iteration scheme
\begin{equation}\label{iter_1}
  \left\{
  \begin{array}{lll}
\mu^{\hat{t}}&=&\displaystyle\biggl.\frac{-\lambda\displaystyle\biggl.\int_{\Omega}\frac{ {z}^{\hat{t}}(x)}{\sqrt{d(x)}}\left(\hat{k}_{\epsilon}*\left(\triangle_{w}(k_{\epsilon}*\frac{{z}^{\hat{t}}}{\sqrt{d}})\right)\right)(x)dx-\eta\displaystyle\biggl.\int_{\Omega}\frac{ {z}^{\hat{t}}(x)}{\sqrt{d(x)}}\triangle\frac{{z}^{\hat{t}}(x)}{\sqrt{d(x)}}dx}{\displaystyle\biggl.\int_{\Omega}(z^{\hat{t}})^{2}(x)dx},\\
\hat{{z}}^{\hat{t}+1}(x)&=&{z}^{\hat{t}}(x)-\tau\left(-\displaystyle\biggl.\frac{\lambda}{\sqrt{d(x)}}\left(\hat{k}_{\epsilon}*\left(\triangle_{w}(k_{\epsilon}*\frac{z^{\hat{t}}}{\sqrt{d}})\right)\right)(x)-\frac{\eta}{\sqrt{d(x)}}\triangle\frac{z^{\hat{t}}(x)}{\sqrt{d(x)}}-\mu^{\hat{t}} z^{\hat{t}}(x)\right),\\
~&~\\
{z}^{\hat{t}+1}(x)&=&Proj_{\mathbb{S}_{2}}(\hat{{z}}^{\hat{t}+1}(x)),\\
  \end{array}
  \right.
\end{equation}
where $\hat{t}=0,\cdots,\hat{T}$ is an inner iteration to have an approximation solution of (\ref{dinkelbach_1}). In fact, in our problem, the Dinkelbach scheme is essentially the Lagrangian method in which the multiplier $\mu$ is given by the fractional-form iteration, which makes the inner iteration (\ref{iter_1}) be convergent \cite{dncut1} and  stable.
As a numerical verification, we show the variation of the Lagrangian multiplier $\mu$ during the iteration in figure \ref{fig:5}.


%
%
%

{
Here, we list the formulation of projection
\begin{equation*}
{z}^{\hat{t}+1}(x)=Proj_{\mathbb{S}_{2}}(\hat{{z}}^{\hat{t}+1}(x)).
\end{equation*}
It equals to the saddle point problem below
\begin{equation*}
\min\limits_{z}\max\limits_{\pi}\frac{1}{2}\displaystyle\biggl.\int_{\Omega}\left({z}(x)-\hat{{z}}^{\hat{t}+1}(x)\right)^{2}dx
-\pi\displaystyle\biggl.\int_{\Omega}\sqrt{d(x)}z(x)dx,
\end{equation*}
where $\pi$ is the Lagrangian multiplier.}

{
Then according to the optimal condition, we have
\begin{equation}\label{project_1}
{z}^{\hat{t}+1}(x)=\hat{{z}}^{\hat{t}+1}(x)+\pi \sqrt{d(x)}.
\end{equation}
Since ${z}^{\hat{t}+1}\in\mathbb{S}_{2}$, we multiply the term $\sqrt{d(x)}$ on both sides of (\ref{project_1}), and then take integration over $\Omega$, we have
\begin{equation}\label{project_2}
\pi=-\frac{\displaystyle\biggl.\int_{\Omega}\hat{{z}}^{\hat{t}+1}(x)\sqrt{d(x)}dx}{\displaystyle\biggl.\int_{\Omega}d(x)dx}.
\end{equation}
Plugging (\ref{project_2}) into (\ref{project_1}), then
$$
{z}^{\hat{t}+1}(x)=Proj_{\mathbb{S}_{2}}(\hat{{z}}^{\hat{t}+1}(x))=\hat{{z}}^{\hat{t}+1}(x)
-\frac{\displaystyle\biggl.\int_{\Omega}\hat{{z}}^{\hat{t}+1}(x)\sqrt{d(x)}dx}{\displaystyle\biggl.\int_{\Omega}d(x)dx}\sqrt{d(x)}.
$$
Once we get the solution of $z$, then we can easily recover ${f}$ by
$f(x)=\frac{z(x)}{\sqrt{d(x)}}$.
}

In summary, we give the algorithm 1 for NCAS$\text{H}^{1}$ Model.
\begin{algorithm}\label{alg_1}
	\caption{NCAS$\text{H}^{1}$ Model}
1.Given ${f}^{0}={1}$, a tolerant error = $\epsilon$. Set $\tau$=2 ,$h^{0}$=50, $\hat{T}=1000$. Let $t=0$.\\
\\
2.Update SMF
$$
w^{t+1}(x,y)=\frac{e^{\frac{-(I(x)-I(y))^{2}}{2(h^{t})^{2}}-\lambda [(k_{\epsilon}*f^{t})(x)-(k_{\epsilon}*f^{t})(y)]^{2}}}{\displaystyle\biggl.\int_{\Omega}e^{\frac{-(I(x)-I(y))^{2}}{2(h^{t})^{2}}-\lambda [(k_{\epsilon}*f^{t})(x)-(k_{\epsilon}*f^{t})(y)]^{2}}dy}.
$$
3.Project $w^{t+1}$ onto $\mathbb{C}_{2}$
$$
w^{t+1}(x,y):=\frac{w^{t+1}(x,y)+w^{t+1}(y,x)}{2}.
$$
4. Calculate $d(x)=\int_{\Omega}w^{t+1}(x,y) dy$.\\
5. Update $h$
\begin{equation*}
(h^{t+1})^{2}=Proj_{\mathbb{H}}(\frac{\displaystyle\biggl.\int_{\Omega\times\Omega}w^{t+1}(x,y)[I(x)-I(y)]^{2}dxdy}{|\Omega|}).
\end{equation*}
\\
6. Let ${z}^{t,0}=\sqrt{d}{f}^{t}$, and calculate $z^{t}$ with an inner iteration $\hat{t}=0,1,2,...,\hat{T}$
\begin{equation*}
  \left\{
  \begin{array}{lll}
\mu^{t,\hat{t}}&=&\displaystyle\biggl.\frac{-\lambda\displaystyle\biggl.\int_{\Omega}\frac{ {z}^{t,\hat{t}}(x)}{\sqrt{d(x)}}\left(\hat{k}_{\epsilon}*\left(\triangle_{w^{t+1}}(k_{\epsilon}*\frac{{z}^{t,\hat{t}}}{\sqrt{d}})\right)\right)(x)dx-\eta\displaystyle\biggl.\int_{\Omega}\frac{ {z}^{t,\hat{t}}(x)}{\sqrt{d(x)}}\triangle\frac{{z}^{t,\hat{t}}(x)}{\sqrt{d(x)}}dx}{\displaystyle\biggl.\int_{\Omega}(z^{t,\hat{t}})^{2}(x)dx},\\
\hat{{z}}^{t,\hat{t}+1}(x)&=&{z}^{t,\hat{t}}(x)-\tau\left(-\displaystyle\biggl.\frac{\lambda}{\sqrt{d(x)}}\left(\hat{k}_{\epsilon}*\left(\triangle_{w^{t+1}}(k_{\epsilon}*\frac{z^{t,\hat{t}}}{\sqrt{d}})\right)\right)(x)-\frac{\eta}{\sqrt{d(x)}}\triangle\frac{z^{t,\hat{t}}(x)}{\sqrt{d(x)}}-\mu^{t,\hat{t}} z^{t,\hat{t}}(x)\right),\\
~&~\\
{z}^{t,\hat{t}+1}(x)&=&Proj_{\mathbb{S}_{2}}(\hat{{z}}^{t,\hat{t}+1})(x),\\
  \end{array}
  \right.
\end{equation*}
7. Let ${z}^{t+1}(x)={z}^{t,\hat{T}}(x)$, and reconstruct segmentation vector
$$
{f}^{t+1}(x)=\frac{{z}^{t+1}(x)}{\sqrt{d(x)}}.
$$
8.If $\frac{||{f}^{t+1}-{f}^{t}||^{2}}{||{f}^{t}||^{2}}<\epsilon$, stop; else, set $t=t+1$, go to step 2.
\end{algorithm}

\subsection{Algorithm for NCASTV Model}
\label{sec_4.2}
Similarly, the NCASTV model  can also be directly minimized by alternating minimization algorithm, one may have these subproblems
\begin{equation*}
\left\{
   \renewcommand\arraystretch{2}{
  \begin{array}{rll}
\min\limits_{w\in \mathbb{C}_{2}}\max\limits_{\beta}&\left\{\displaystyle\biggl.\int_{\Omega\times\Omega}\left\{\frac{(I(x)-I(y))^{2}}{2h^{2}}+\ln(\sqrt{2\pi}h|\Omega|)\right\}w(x,y)dxdy\right.
+\displaystyle\biggl.\int_{\Omega\times\Omega}w(x,y)\ln w(x,y)dxdy\\
&+\displaystyle\biggl.\int_{\Omega}\beta(x)(1-\displaystyle\biggl.\int_{\Omega}w(x,y)dy)dx
+\left.\lambda\displaystyle\biggl.\int_{\Omega\times\Omega} w(x,y)[(k_{\epsilon}*f)(x)-(k_{\epsilon}*f)(y)]^{2}dxdy\right\},\\
\min\limits_{h\in\mathbb{H}}& \left\{\displaystyle\biggl.\int_{\Omega\times\Omega}\left\{\frac{(I(x)-I(y))^{2}}{2h^{2}}+\ln(\sqrt{2\pi}h|\Omega|)\right\}w(x,y)dxdy\right\},\\
\min\limits_{f\in\mathbb{F}}&\left\{\lambda\displaystyle\biggl.\int_{\Omega\times\Omega}w(x,y)((k_{\epsilon}*f)(x)-(k_{\epsilon}*f)(y))^{2}dxdy
+\eta\displaystyle\biggl.\int_{\Omega} ||\nabla f(x)||dx\right\},\\
     \end{array}}
\right.
\end{equation*}
where $\beta$ is the Lagrangian multiplier, and $\lambda$ and $\eta$ are the parameters.

The $w$ and $h$ subproblems are the same as the NCAS$\text{H}^{1}$ model. Thus, we only need to address the $f$-subproblem
\begin{equation}
\min\limits_{f\in\mathbb{F}}\left\{\lambda\displaystyle\biggl.\int_{\Omega\times\Omega}w(x,y)((k_{\epsilon}*f)(x)-(k_{\epsilon}*f)(y))^{2}dxdy
+\eta\displaystyle\biggl.\int_{\Omega} ||\nabla f(x)||dx\right\}.
\end{equation}

{To solve this problem, we adopt the splitting and penalty methods. It can also be solved by many other splitting methods such as  split-Bregman \cite{sbm}, augmented Lagrangian multiplier method \cite{29} \emph{etc.}. Here, we introduce an auxiliary function ${g}$, which satisfies ${g}={f}$, then we have the following approximation problem
\begin{equation}\label{tv_f}
\min\limits_{g,f\in\mathbb{F}}\left\{\lambda\displaystyle\biggl.\int_{\Omega\times\Omega}w(x,y)((k_{\epsilon}*f)(x)-(k_{\epsilon}*f)(y))^{2}dxdy
+\eta\displaystyle\biggl.\int_{\Omega} ||\nabla g(x)||dx+\epsilon||{f}-{g}||^{2}\right\},
\end{equation}
where $\epsilon$ is a penalty parameter.}

{Thus, we have two subproblems of problem (\ref{tv_f}),
\begin{subequations}\label{tv_f1}
\begin{align}
    \min\limits_{f\in\mathbb{F}}&\left\{\lambda\displaystyle\biggl.\int_{\Omega\times\Omega}w(x,y)((k_{\epsilon}*f)(x)-(k_{\epsilon}*f)(y))^{2}dxdy+\epsilon||{f}-{g}||^{2}\right\},\label{tv_f1_a}\\
    \min\limits_{{g}}&\left\{\eta\displaystyle\biggl.\int_{\Omega} ||\nabla g(x)||dx+\epsilon||{f}-{g}||^{2}\right\}.\label{tv_f1_b}
\end{align}
\end{subequations}}

As we can see, the subproblem (\ref{tv_f1_b}) is the standard ROF model \cite{rof} for denoising, and the subproblem (\ref{tv_f1_a}) is a linear problem which is similar to problem (\ref{h1_f}). So our model can be regarded as an alternating process of Ncut clustering and denoising of the results. It is reasonable that our model will have a better performance under noise. The ROF model can be efficiently solved by many methods such as \cite{7,13ca,29}, here, we choose augmented Lagrangian method \cite{29} to solve it.
As for the subproblem (\ref{tv_f1_a}), please notice that it is a linear problem other than an eigenvalue problem, which corresponds to a fractional programming other than Rayleigh quotient formula. We can also adopt Dinkelbach method \cite{dncut1,dncut2}.

{Denote $z(x)=\sqrt{d(x)}f(x)$, the $f$-subproblem can be converted to the problem of $z$:
\begin{equation}\label{tv_f2}
\min_{\scriptstyle \int_{\Omega}z^2(x) dx=1\atop \scriptstyle \int_{\Omega}\sqrt{d(x)}z(x) dx =0}\left\{\lambda\displaystyle\biggl.\int_{\Omega\times\Omega}w(x,y)((k_{\epsilon}*\frac{z}{\sqrt{d}})(x)-(k_{\epsilon}*\frac{z}{\sqrt{d}})(y))^{2}dxdy
+\epsilon||\frac{z}{\sqrt{d}}-{g}||^{2}\right\}.
\end{equation}}

Here we adopt the same procedure as the NCAS$\text{H}^{1}$ model. We calculate the optimal point in $\mathbb{S}_{1}$ by Lagrangian multiplier method, and then project the optimal point onto $\mathbb{S}_{2}$. The optimization problem (\ref{tv_f2}) becomes
\begin{equation}\label{dinkelbach_2}
\begin{array}{lll}
\min\limits_{\textbf{z}\in\mathbb{S}_{2}}\max\limits_{\mu}& \left\{
\lambda\displaystyle\biggl.\int_{\Omega\times\Omega}w(x,y)((k_{\epsilon}*\frac{z}{\sqrt{d}})(x)-(k_{\epsilon}*\frac{z}{\sqrt{d}})(y))^{2}dxdy
+\epsilon||\frac{z}{\sqrt{d}}-{g}||^{2}+\mu\left(1-\int_{\Omega}z^2(x) dx\right)\right\}.
\end{array}
\end{equation}
To solve the above problem, by the first order optimal condition of ${z}$, we have
\begin{equation}\label{tvf_1}
-\frac{\lambda}{\sqrt{d(x)}}\left(\hat{k}_{\epsilon}*\left(\triangle_{w}(k_{\epsilon}*\frac{z}{\sqrt{d}})\right)\right)(x)
+\epsilon\left(\frac{z(x)}{d(x)}-\frac{g(x)}{\sqrt{d(x)}}\right)-\mu{z(x)}=0,
\end{equation}
then
\begin{equation}
\mu=\displaystyle\biggl.\frac{-\lambda\displaystyle\biggl.\int_{\Omega}\frac{ {z}(x)}{\sqrt{d(x)}}\left(\hat{k}_{\epsilon}*\left(\triangle_{w}(k_{\epsilon}*\frac{{z}}{\sqrt{d}})\right)\right)(x)dx
+\displaystyle\biggl.\int_{\Omega}\epsilon\left(\frac{z^{2}(x)}{d(x)}-\frac{z(x)g(x)}{\sqrt{d(x)}}\right)dx}
{\displaystyle\biggl.\int_{\Omega}z^{2}(x)dx}.
\end{equation}

Following the previous discussion, we construct the iteration scheme below,
{
\begin{equation}\label{iter_2}
  \left\{
  \begin{array}{lll}
\mu^{\hat{t}}&=&\displaystyle\biggl.\frac{-\lambda\displaystyle\biggl.\int_{\Omega}\frac{ {z}^{\hat{t}}(x)}{\sqrt{d(x)}}\left(\hat{k}_{\epsilon}*\left(\triangle_{w}(k_{\epsilon}*\frac{{z}^{\hat{t}}}{\sqrt{d}})\right)\right)(x)dx
+\displaystyle\biggl.\int_{\Omega}\epsilon\left(\frac{(z^{\hat{t}})^{2}(x)}{d(x)}-\frac{z^{\hat{t}}(x)g(x)}{\sqrt{d(x)}}\right)dx}
{\displaystyle\biggl.\int_{\Omega}(z^{\hat{t}})^{2}(x)dx},\\
\hat{{z}}^{\hat{t}+1}(x)&=&{z}^{\hat{t}}(x)-\tau\left(
-\frac{\lambda}{\sqrt{d(x)}}\left(\hat{k}_{\epsilon}*\left(\triangle_{w}(k_{\epsilon}*\frac{z^{\hat{t}}}{\sqrt{d}})\right)\right)(x)
+\epsilon\left(\frac{z^{\hat{t}}(x)}{d(x)}-\frac{g(x)}{\sqrt{d(x)}}\right)-\mu^{\hat{t}}{z^{\hat{t}}(x)}
\right),\\
~&~\\
{z}^{\hat{t}+1}(x)&=&Proj_{\mathbb{S}_{2}}(\hat{{z}}^{\hat{t}+1}(x))=\hat{{z}}^{\hat{t}+1}(x)
-\frac{\displaystyle\biggl.\int_{\Omega}\hat{{z}}^{\hat{t}+1}(x)\sqrt{d(x)}dx}{\displaystyle\biggl.\int_{\Omega}d(x)dx}\sqrt{d(x)}.\\
  \end{array}
  \right.
\end{equation}
}

Then we can recover ${f}$ by
$f(x)=\frac{z(x)}{\sqrt{d(x)}}$. Finally, we summarize the algorithm for NCASTV in algorithm 2.


\begin{algorithm}\label{alg2}
	\caption{NCASTV Model}
1. Given ${f}^{0}={g}^{0}={1}$, tolerant error = $\zeta$. Set $\tau$=2, $h^{0}=50$, $\hat{T}=1000$. Let $t=0$.\\
\\
2. Update SMF
$$
w^{t+1}(x,y)=\frac{e^{\frac{-(I(x)-I(y))^{2}}{2(h^{t})^{2}}-\lambda [(k_{\epsilon}*f^{t})(x)-(k_{\epsilon}*f^{t})(y)]^{2}}}{\displaystyle\biggl.\int_{\Omega}e^{\frac{-(I(x)-I(y))^{2}}{2(h^{t})^{2}}-\lambda [(k_{\epsilon}*f^{t})(x)-(k_{\epsilon}*f^{t})(y)]^{2}}dy}.
$$
3. Project $w^{t+1}$ onto $\mathbb{C}_{2}$
$$
w^{t+1}(x,y):=\frac{w^{t+1}(x,y)+w^{t+1}(y,x)}{2}.
$$
4. Calculate $d(x)=\int_{\Omega}w^{t+1}(x,y) dy$.\\

5.Calculate $h$
\begin{equation*}
(h^{t+1})^{2}=Proj_{\mathbb{H}}(\frac{\displaystyle\biggl.\int_{\Omega\times\Omega}w^{t+1}(x,y)[I(x)-I(y)]^{2}dxdy}{|\Omega|}).
\end{equation*}
\\
6. Let ${z}^{t,0}=\sqrt{d}{f}^{t}$, and calculate ${z}$ with inner loop iteration $\hat{t}=0,1,..,\hat{T}$
\begin{equation*}
  \left\{
  \begin{array}{lll}
\mu^{t,\hat{t}}&=&\displaystyle\biggl.\frac{-\lambda\displaystyle\biggl.\int_{\Omega}\frac{ {z}^{t,\hat{t}}(x)}{\sqrt{d(x)}}\left(\hat{k}_{\epsilon}*\left(\triangle_{w^{t+1}}(k_{\epsilon}*\frac{{z}^{t,\hat{t}}}{\sqrt{d}})\right)\right)(x)dx
+\displaystyle\biggl.\int_{\Omega}\epsilon\left(\frac{(z^{t,\hat{t}})^{2}(x)}{d(x)}-\frac{z^{t,\hat{t}}(x)g^{t}(x)}{\sqrt{d(x)}}\right)dx}
{\displaystyle\biggl.\int_{\Omega}(z^{t,\hat{t}})^{2}(x)dx},\\
\hat{{z}}^{t,\hat{t}+1}(x)&=&{z}^{t,\hat{t}}(x)-\tau\left(
-\frac{\lambda}{\sqrt{d(x)}}\left(\hat{k}_{\epsilon}*\left(\triangle_{w^{t+1}}(k_{\epsilon}*\frac{z^{t,\hat{t}}}{\sqrt{d}})\right)\right)(x)
+\epsilon\left(\frac{z^{t,\hat{t}}(x)}{d(x)}-\frac{g^{t}(x)}{\sqrt{d(x)}}\right)-\mu^{t,\hat{t}}{z^{t,\hat{t}}(x)}
\right),\\
~&~\\
{z}^{t,\hat{t}+1}(x)&=&Proj_{\mathbb{S}_{2}}(\hat{{z}}^{t,\hat{t}+1}(x))=\hat{{z}}^{t,\hat{t}+1}(x)
-\frac{\displaystyle\biggl.\int_{\Omega}\hat{{z}}^{t,\hat{t}+1}(x)\sqrt{d(x)}dx}{\displaystyle\biggl.\int_{\Omega}d(x)dx}\sqrt{d(x)}.\\
  \end{array}
  \right.
\end{equation*}
7. Let ${z}^{t+1}={z}^{t,\hat{T}}$, and reconstruct ${f}$
$$
{f}^{t+1}(x)=\frac{{z}^{t+1}(x)}{\sqrt{d(x)}}.
$$
8. Calculate the auxiliary variable
$${g}^{t+1}=ROF({f}^{t+1},\frac{\eta}{2\epsilon}).$$
9. If $\mathnormal{\frac{||{f}^{t+1}-{f}^{t}||^{2}}{||{f}^{t}||^{2}}}<\zeta$, stop; else, set $t=t+1$, return to step 2.
\end{algorithm}

\section{Experimental Results}
\label{sec_5}
The main contribution of the proposed models are the adaptivity of SMF together with a spatial regularization. Here we design several experiments to show the function of these two aspects. After that,  we will compare our models with some classical segmentation models, such as the classical Chan-Vese model and the original Ncut-based segmentation model. To reduce the computational cost, in the following experiments, we set $k_{\epsilon}$ to be the delta function $\delta$, and thus $k_{\epsilon}*f=f$.

{Let us point out that we have to compute a similarity matrix which has $O(|\Omega|^2)$ complexity in each outer iteration if one would like to use a fully connected similarity matrix. This is a very large computational burden. Thus in real implementation, we just compute and store the most $k$ related and important values for $w$. In fact, there are two main techniques can be adopted to save storage and speed up the algorithm in spectral clustering. One is the thresholding technique, and the other is the k-nearest neighbors method. In this paper, we adopt k-nearest neighbors whose size is $21 \times 21$.}

In the following experiments, all the natural images are taken from BSDS500 database\footnote{BSDS500 database: https://www2.eecs.berkeley.edu/Research/Projects/CS/vision/grouping/resources.html} \cite{bsds}.

\subsection{Toy Experiments}
\label{sec_5.0}

{To compare with the original Ncut model, we take a classical example to show the improvements of our method. In this toy data, 300 points in
$\mathbb{R}^2$ form a double-moon shape. These points are hoped to be separated into two classes in which one is the the upper half moon and the other is the lower half moon. Let us emphasize that this data is not centrally separable, thus the center-based segmentation algorithms such as K-means and CV model could not finish this task well. In this experiment, we take $I(x)$ as the coordinates of each point.
As for the noise, it implies the perturbation of the coordinates of data.}

To see the role of the adaptive similarity measure, we design two experiments.

\begin{figure*}
\includegraphics[width=0.9\textwidth]{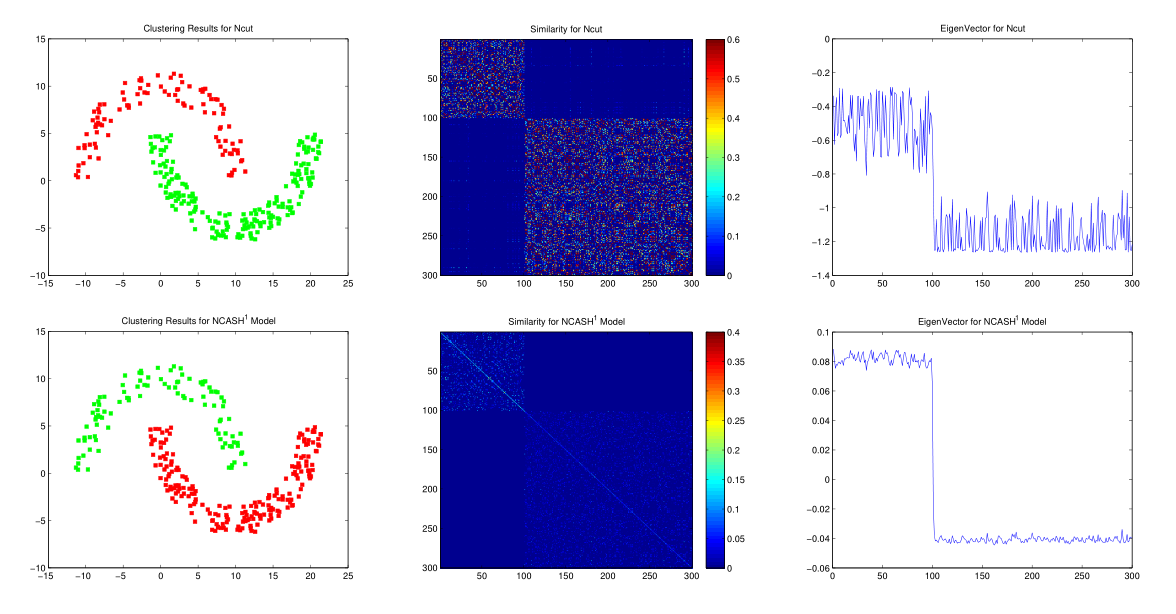}
\caption{Results of Ncut model (1st row) and the proposed NCAS\text{H}$^{1}$ (2nd row) for noise free double-moon data. The related similarity matrices and eigenvectors are shown in the last two columns.}
\label{fig:3}
\end{figure*}

{
In the first experiment, we test Ncut and our algorithm on a clean double-moon data. Let us first give some parameters appeared in Ncut and ours. For Ncut algorithm, the similarity is given by $w(x,y)=e^{\frac{-(I(x)-I(y))^{2}}{h^{2}}}$ with $h=3$. While in our NCAS$\text{H}^{1}$, the
parameter $\lambda=1$ and $\eta=0.25*\lambda$. The
clustering results of two methods are shown in the first column (\figurename~\ref{fig:3}). The next two columns contain the related similarity matrices and computed eigenvectors, respectively. It is easy to find out that both of the two methods can produce good clustering results in this noise free case. By carefully observing these two similarity matrices, the similarity with adaptive $h$ in our algorithm can produce
more binary connections than Ncut's. As can be found in this figure, the left lower and upper right of $w$ in NCAS$\text{H}^{1}$ are almost $0$, which implies that these points have less similarities. But in Nuct's, there are some values of $w$ more than $0$ in the same regions, which means there may be some dependences among these points. If there are some noises, this will lead to some misclassifications.
Besides, the eigenvectors produced by Ncut is oscillating, though it can classify the data correctly in this noise free case, it would fail when the data corrupted by heavy noise. On the other hand, the eigenvector $f$ provided by the proposed NCAS$\text{H}^{1}$ is smooth, and one can easily get two latent classes according to a simple threshold value.
}

\begin{figure*}
\includegraphics[width=0.9\textwidth]{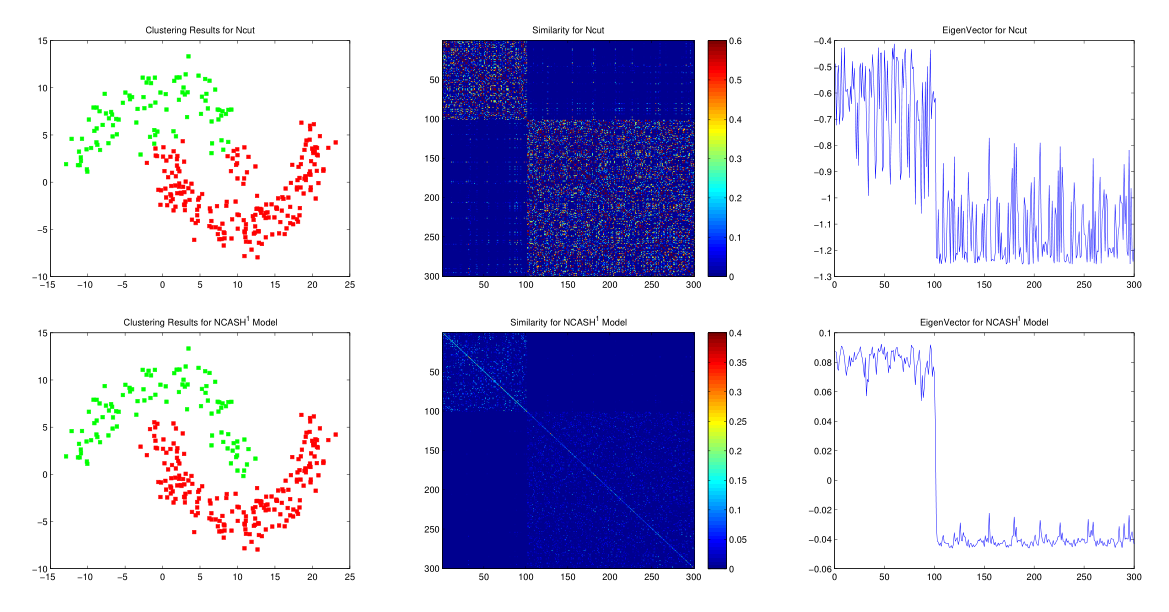}
\caption{Results of Ncut model (1st row) and the proposed NCAS\text{H}$^{1}$ (2nd row) for double-moon data corrupted by Gaussian noise. The related similarity matrices and eigenvectors are shown in the last two columns.}
\label{fig:4}
\end{figure*}

{In the second experiment (\figurename~\ref{fig:4}), we test them on the double-moon data set corrupted by Gaussian noise with distribution N(0,1).
In this case, Ncut model produces undesirable results, which contains 24 wrong-labeled points. But our NCAS$\text{H}^{1}$ model can partition all the data points correctly due to the existence of regularizer and updating similarity. Since the adaptive similarity plays a vital role in our model, as before, we show the similarity matrices of Ncut model and NCAS$\text{H}^{1}$. One can find out that the similarity produced by our method is nearly block-diagonal, which indicates a clear connection and is beneficial for a clustering process. In this noisy case, the similarity of the Ncut model is not so ``clean" under noise, which means some relationships of data are mistaken. As for the eigenvectors used for clustering, in the noisy data, the eigenvector by the Ncut model has serious oscillations and it fails to provide a good clustering criterion. On the contrary, the eigenvector calculated by the proposed NCASH$^{1}$ model still has a big jump and less oscillations. This two experiments show that the proposed method is more robust than the Ncut model.}


\begin{figure*}
\includegraphics[width=0.9\textwidth]{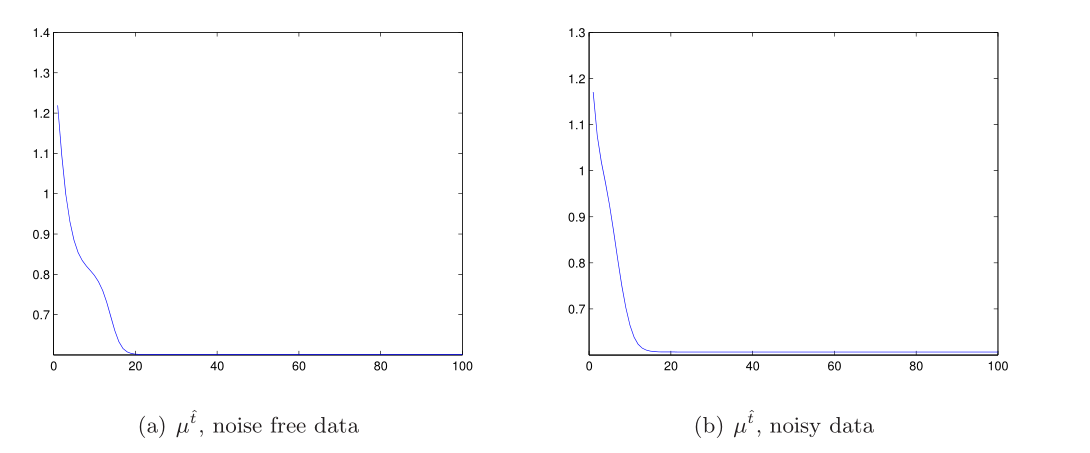}
\caption{The values of $\mu^{\hat{t}}$ (the $10$-th outer iteration) appeared in $NCASH^{1}$ model for the noise data and noisy data, respectively.}
  \label{fig:5}
\end{figure*}

{ It seems that the $\mu^{\hat{t}}$ in \eqref{iter_1} is not increasing during the inner iteration, as a numerical verification, we display the $\mu^{\hat{t}}$'s values during the first 100 inner iterations in \figurename~\ref{fig:5}, which show that $\mu^{\hat{t}}$ is convergent numerically.}

In the next, we will show the performance of the updating similarity measure and regularization for segmentation of natural images.
\subsection{Performance of Regularization}
\label{sec_5.1}

\begin{figure*}
\includegraphics[width=0.9\textwidth]{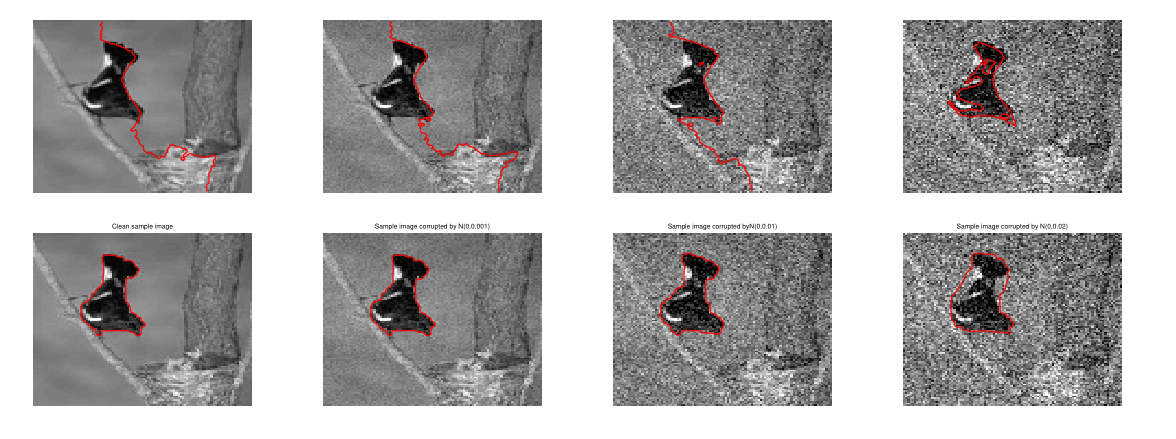}
\caption{The comparison of Ncut (the first row) and NCASTV (the second row) under different levels of Gaussian noise. }
  \label{fig:10}
\end{figure*}

The introduction of regularization makes our  models more robust to noise. Here two experiments are designed to show the effect of regularization in our proposed model. The first experiment is designed to show the function of regularization under different levels of noise. In this experiment, we compare Ncut-based segmentation model with our proposed NCASTV model. We show the segmentation results of images corrupted by Gaussian noise with different levels: N(0,0), N(0,0.001), N(0,0.01), N(0,0.02), respectively in \figurename~\ref{fig:10}. And the parameters in NCASTV model are set as: $\lambda=1$, $\epsilon=0.001*\lambda$, $\eta=0.005*\epsilon$ under noise $N(0,0), N(0,0.001),N(0,0.01),$ and $\eta=0.009*\epsilon$ under noise $N(0,0.02)$. Compared to Ncut, the experiments (\figurename~\ref{fig:10}) show that our model is robust under noise with different levels, since the spatial priori (TV regularization) plays a vital role in our model.

\begin{figure*}
\includegraphics[width=0.9\textwidth]{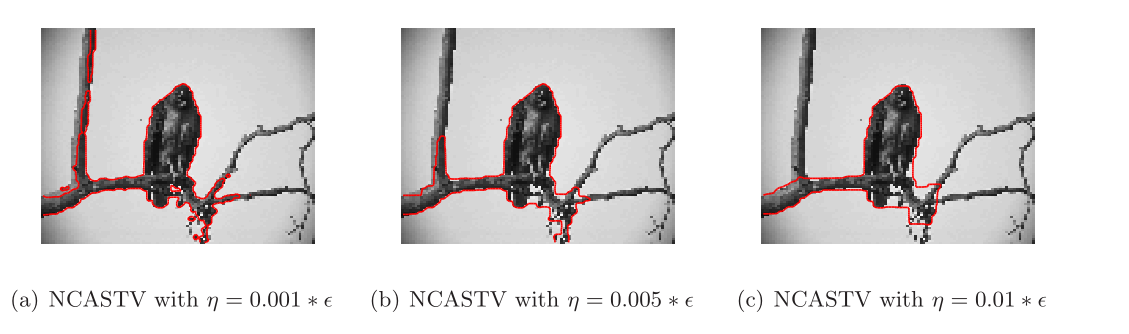}
\caption{The segmentation results of NCASTV model with different regularization parameters.}
  \label{fig:12}
\end{figure*}

 {In the next experiment, we simply set the regularization parameter $\eta$ as different values: $0.001*\epsilon,~0.005*\epsilon,~0.01*\epsilon$, and other parameters are set as: $\lambda=1$, $\epsilon=0.001*\lambda$. The results in \figurename~\ref{fig:12} show that the segmentation results of the sample image become more smooth and the lengths of contours become shorter as the regularization parameters $\eta$ become bigger.} Both of these two experiments demonstrate the function of the regularization in our models.

\subsection{The Performance of Adaptive Similarity}

\begin{figure*}
\includegraphics[width=0.9\textwidth]{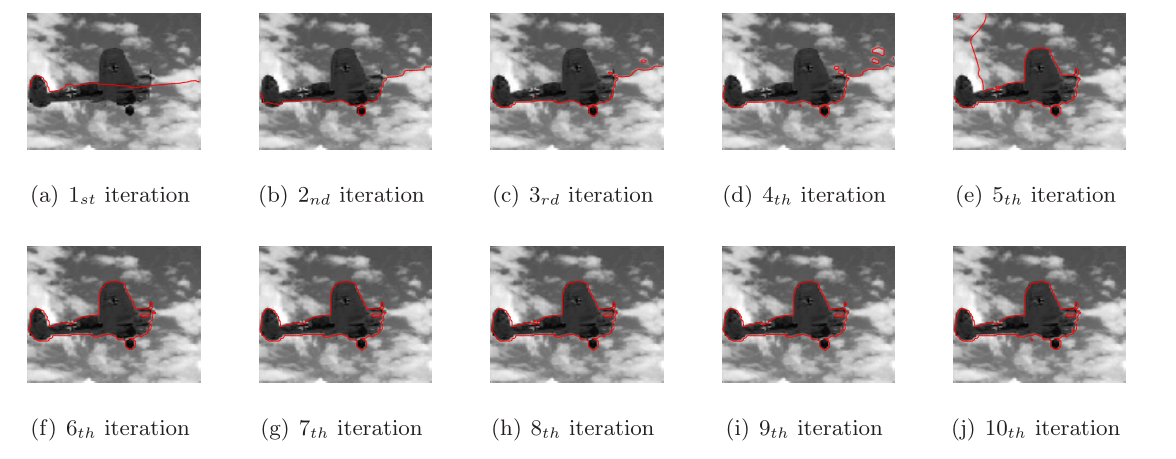}
\caption{(a)-(j) show the segmentation results of the first 10 iterations of an image by NCASTV model. }
  \label{fig:9}
\end{figure*}

To be contrasted with the traditional Ncut model, the similarity in our proposed models is determined by the energy functional itself, and it can be updated during the iteration. Here we pay attention to NCASTV model and demonstrate the contribution of similarity updating by showing the details of the iterations in the experiments below. The parameters in the NCASTV model are set as: $\lambda=1$, $\epsilon=0.001*\lambda$, $\eta=0.001*\epsilon$, and $h^{0}=50$. We show the results of first 10 iterations by NCASTV model in (\figurename~\ref{fig:9}).

{From this experiment, we can find out that the segmentation results are greatly improved with the on-going iterations, since the similarity is adaptively updated by the model to fit the data better. In fact, we establish a better classification criterion compared with the traditional Ncut-based model \cite{3sm}.}

\subsection{Comparisons among Chan-Vese, Pre-Ncut, NCAS$\text{H}^{1}$ and the NCASTV model}
\label{sec_5.2}
{Since there is no spatial prior information for the segmentation results by the Ncut model, the segmentation results are always undesirable under noise.} To improve the performance of Ncut-based model, a preprocessing technique is applied in Ncut-based model \cite{3sm}. It uses a kernel-based filter to generate an edge-based image, and the similarity of that image is calculated by
\begin{equation*}
w(x,y)
=\left\{
  \begin{array}{ll}
    e^{-\frac{(||(\nabla G*I)(x)||_{2}-||(\nabla G*I)(y)||_{2})^{2}}{2h^{2}}}&,x\neq y,\\
    1&,x=y.\\
  \end{array}
  \right.
\end{equation*}
where $G$ is the filter kernel. We denote this algorithm as Pre-Ncut model.
In fact, Pre-Ncut is an edge-based segmentation method, and it highly depends on the edge detectors.

\begin{figure*}
\includegraphics[width=0.9\textwidth]{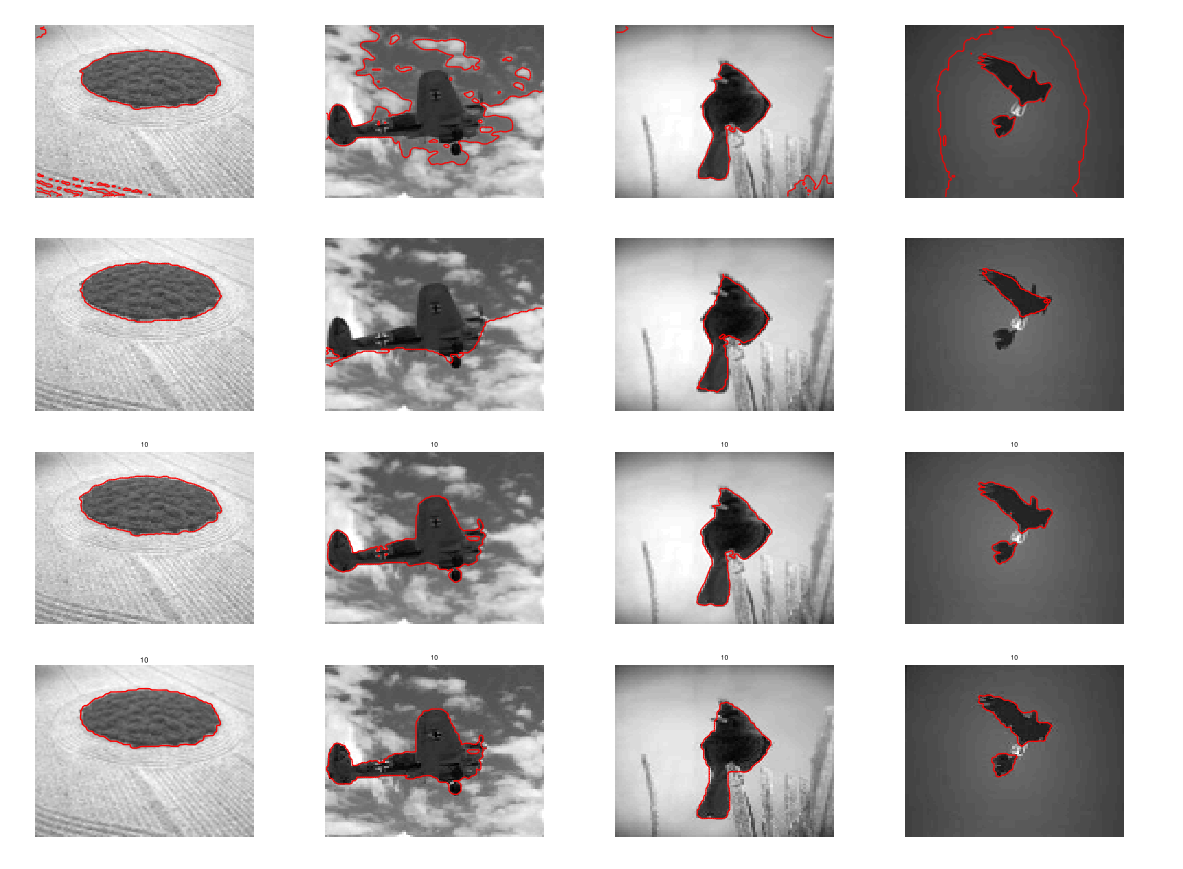}
\caption{Results by Chan-Vese model (1st row), Pre-Ncut (2nd row), NCAS$\text{H}^{1}$ model (3rd row) and NCASTV model (4th row). The parameters in NCASH$^{1}$-based model: $\lambda=1,\eta=0.001*\lambda,0.001*\lambda,0.0005*\lambda,0.001*\lambda$, respectively. The parameters in NCASTV model: $\lambda = 1$, $\epsilon = 0.001*\lambda$, and $\eta=0.001*\epsilon, 0.001*\epsilon, 0.008*\epsilon, 0.003*\epsilon$, respectively.}
  \label{fig:7}
\end{figure*}

In the following experiments, we will provide some comparisons between the Chan-Vese model \cite{1cv}, the Pre-Ncut \cite{3sm} algorithm, and the proposed models. The BSDS500 database is usually used for segmentation evaluation, which consists of natural images and their corresponding ground truth. There are many multiscale ground truth in this data set. We give a binary ground truth by merging some classes to calculate the segmentation accuracy since the proposed models are used for two-phase segmentation. Besides, to test our method efficiently and save memory storage, we resize the original images and the corresponding ground truth to the size of $100\times100$. Here we choose four images from BSDS500 database which are denoted as Image1, Image2, Image3,  and Image4 for convenience. The segmentation results of these methods are shown in \figurename~\ref{fig:7}. In these experiments, we set the parameters in NCASH$^{1}$ model as: $\lambda=1$, and $\eta=0.001,0.001,0.0005,0.001$ for different images, respectively, and parameters in NCASTV model: $\lambda = 1$, $\epsilon = 0.001*\lambda$, and $\eta=0.001*\epsilon, 0.001*\epsilon, 0.008*\epsilon, 0.003*\epsilon$ for different images, respectively.

\subsection*{Evaluation and Analysis}

To evaluate the results obtained by different methods, here we consider two indexes \cite{bsds}.

\begin{enumerate}
\item Variation of information.\\
The Variation of Information (VI) \cite{bsds6} metric is used for clustering comparison, which measures the distance between two clusterings with respect to their average conditional entropy given by
\begin{equation*}\label{vi}
VI(S,S')=H(S)+H(S')-2I(S,S'),
\end{equation*}
where $H$ represents the entropy and $I$ is the mutual information between two clusterings $S$ and $S'$ of data.
\item Rand index.\\
The Rand Index (RI) \cite{bsds62} is designed for clustering evaluation, which measures the similarity between two data clusterings. The RI between test segmentation $S$ and the corresponding ground truth segmentation $G$ is defined as the sum of the amount of pixels pairs with the same labels in $S$ and $G$ and those with different labels in all segmentations, and then divided by the number of pixels pairs \cite{bsds}. Given a test segmentation $S$ and a set of corresponding ground-truth segmentations $\{G_{t}\}$, the RI \cite{bsds5} \cite{bsds7} is given by
\begin{equation}\label{ri}
RI(S,\{G_{t}\})=\frac{1}{T}\sum\limits_{i<j}[c_{ij}p_{ij}+(1-c_{ij})(1-p_{ij})],
\end{equation}
where $c_{ij}$ is the event that pixels $i$ and $j$ with the same label, and the corresponding probability $p_{ij}$. $T$ is the number of pairs of pixels. Here the $p_{ij}$ is estimated by the sample mean, and (\ref{ri}) means to average the RI values of all ground-truth segmentations.
\end{enumerate}

With these two quality indexes, we provide a comparison of the four algorithms: the Chan-Vese model, the Pre-Ncut, the proposed NCAS$\text{H}^{1}$, and the NCASTV models in Table \ref{table_1}.
\begin{table*}[h]
\centering
\caption{Segmentation accuracy of different algorithms.}\label{table_1}
\begin{tabular}{ccccccccc}
\toprule[1pt]
& \multicolumn{2}{c}{Image 1} & \multicolumn{2}{c}{Image 2} &\multicolumn{2}{c}{Image 3}& \multicolumn{2}{c}{Image 4}\\
\cline{2-9}
& VI & RI &VI & RI &VI & RI&VI & RI \\
\noalign{\smallskip}\hline\noalign{\smallskip}
 Chan-Vese&0.2710 &0.9551 &1.2914 &0.5591 &0.3811 &0.9318 &1.2053 &0.5224\\
\noalign{\smallskip}\hline\noalign{\smallskip}
 Pre-Ncut&0.1268 &0.9843 & 1.4567 &0.5026 &0.1760 &0.9736&0.2010 &0.9624 \\
\noalign{\smallskip}\hline\noalign{\smallskip}
 NCAS$\text{H}^{1}$&0.1219 &0.9830 &0.1986 &0.9694 &\textbf{0.1311} &\textbf{0.9827} &\textbf{0.1018} &\textbf{0.9862}\\
\noalign{\smallskip}\hline\noalign{\smallskip}
 NCASTV&\textbf{0.0793} &\textbf{0.9909} &\textbf{0.1930} &\textbf{0.9696} &0.1853 &0.9736 &0.1378 &0.9800\\
\bottomrule[1pt]
\end{tabular}
\end{table*}

\begin{figure*}
\includegraphics[width=0.9\textwidth]{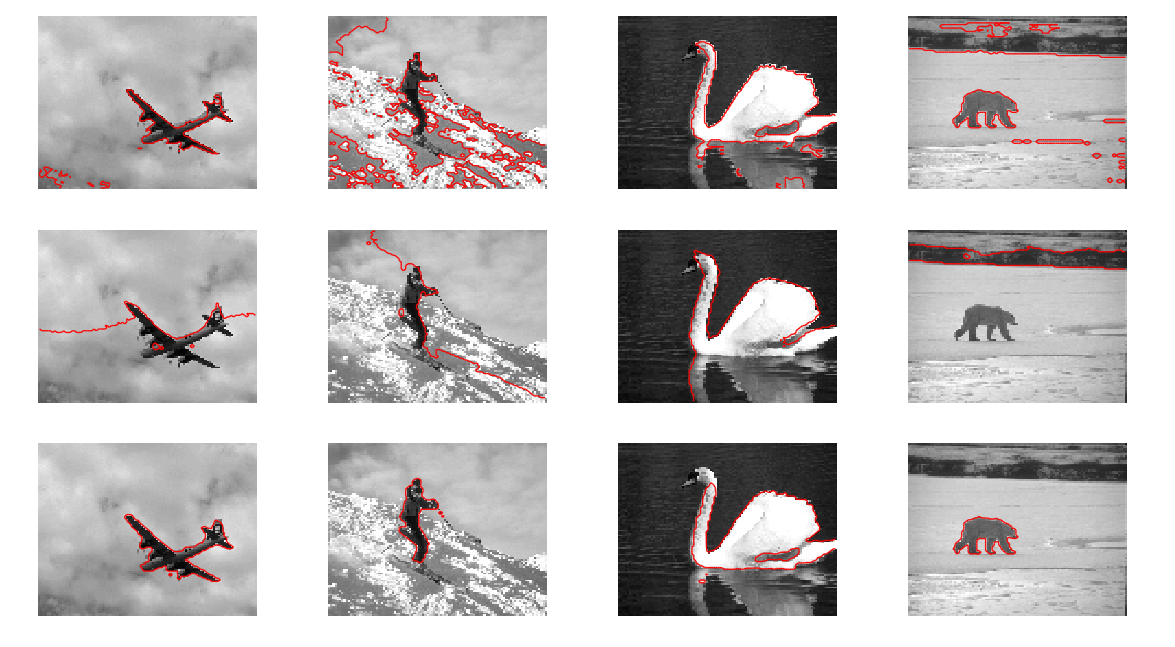}
\caption{Results by Chan-Vese model (1st row), Pre-Ncut (2nd row) and NCASTV model (3rd row), and the parameters in NCASTV model  are: $\lambda = 1$, $\epsilon = 0.001*\lambda$, and $\eta=0.001*\epsilon$ for all images.}
  \label{fig:11}
\end{figure*}

\begin{figure*}
\includegraphics[width=0.9\textwidth]{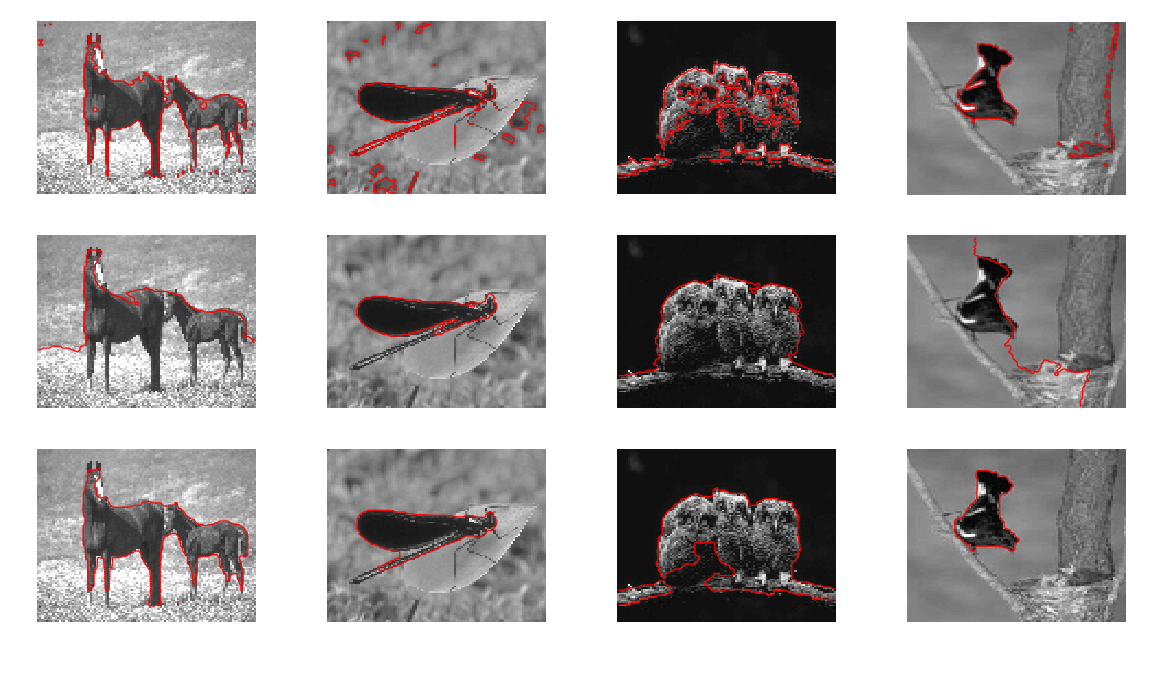}
\caption{{Results by Chan-Vese model (1st row), Pre-Ncut (2nd row) and NCASTV model (3rd row), and parameters in NCASTV model all are: $\lambda = 1$, $\epsilon = 0.001*\lambda$, and $\eta=0.003,0.004,0.005,0.005*\epsilon$.}}
  \label{fig:11_2}
\end{figure*}

The numerical results have shown visually that our proposed models have better performance compared with the traditional Pre-Ncut and the classical  Chan-Vese models. In addition, the quantitative evaluation also demonstrates this conclusion. From Table \ref{table_1}, the results of the proposed models have smaller VI values and larger RI values, which means that the results obtained by proposed models are ``closer" and more similar to the ground truth segmentations than other methods. In \figurename~\ref{fig:11} and \figurename~\ref{fig:11_2}, we give more results produced by different algorithms.

{
There are a few reasons that the proposed models can provide better results.
Firstly, the similarities in our models are determined by the proposed cost functional, which can be adaptively updated to fit the data distribution in a better way;
Secondly, our proposed models have regularization terms which will equip the models with spatial prior information, such that our model would be robust in the presence of noise. Compared with the Chan-Vese model, the essential difference is the data term. The Chan-Vese model is a center-based clustering and our models are spectrum-based algorithm. The spectral clustering can formulate the geometric information of the image data and transfer the image data into a more separable space, which makes Ncut-based model perform better. In fact, the Chan-Vese model can be extended to a likelihood-based variational problem \cite{emref}, and it can be optimized by an EM-type algorithm. However, in that method, the segmentation model is center-based and do not have an adaptive similarity weight. Moreover, our proposed model has a negative entropy term, which can enhance the smoothness of the classification function and make the model more stable compared with binary Chan-Vese type segmentation, especially in numerical algorithm.}\par

\subsection*{Computation Time}

Since our proposed models are Ncut-based, and the similarity is adaptively updated to fit the data, the computation time are much longer than the traditional Ncut model. Here we show the time for each model in Table \ref{table_2}. For both the NCAS$\text{H}^{1}$ model and the NCASTV model, we set $10$ times outer iterations and $\hat{T}=1000$ times inner loop iterations for each outer iteration.
In this paper, we do not focus on the efficiency of the computation time. Designing some fast algorithms could be our future work.

\begin{table*}
\centering
\caption{ CPU Time.}\label{table_2}
\begin{tabular}{ccccc}
\toprule[1pt]
&Image 1 & Image 2 &Image 3 & Image 4 \\
\noalign{\smallskip}\hline\noalign{\smallskip}
 Chan-Vese&18.5524s&20.2881s&18.4705s&18.3743s\\
\noalign{\smallskip}\hline\noalign{\smallskip}
 Pre-Ncut&2.1157s&2.4953s&2.0476s&1.7916s\\
\noalign{\smallskip}\hline\noalign{\smallskip}
 NCAS$\text{H}^{1}$&65.9595s&58.4429s&61.9768s&67.4101s\\
\noalign{\smallskip}\hline\noalign{\smallskip}
 NCASTV&67.7497s&64.5597s&69.8690s&67.3305s\\
\bottomrule[1pt]
\end{tabular}
\end{table*}

\subsection{Pre-Ncut and Pre-NCASTV}
\label{sec_5.3}
Inspired by the technique used in preprocessing Ncut model, we take some precondition to our NCASTV model to establish an edge-based segmentation method. That is, the similarity in this method is adjusted to
\begin{equation*}
w(x,y)
=\left\{
  \begin{array}{ll}
    \frac{e^{-\frac{(||S(x)||_{2}-||S(y)||_{2})^{2}}{2h^{2}}-\lambda (f(x)-f(y))^{2}}}{\displaystyle\biggl.\int_{\Omega}e^{-\frac{(||S(x)||_{2}-||S(y)||_{2})^{2}}{2h^{2}}-\lambda (f(x)-f(y))^{2}}dy}&,x\neq y,\\
    1&,x=y.\\
  \end{array}
  \right.
\end{equation*}
where $S$ is the edge-based image, and $S=(\nabla G*I).$

\begin{figure*}
\includegraphics[width=0.9\textwidth]{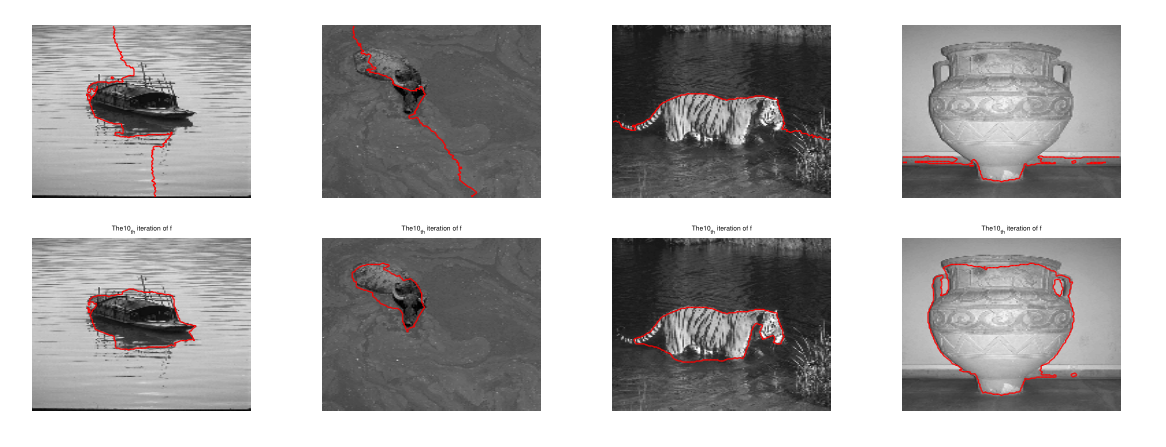}
\caption{ Results by Pre-Ncut model (1st row) and Pre-NCASTV model (2nd row), and the parameters in Pre-NCASTV model: $\lambda = 1$, $\epsilon = 0.001*\lambda$, and $\eta=0.03*\epsilon, 0.005*\epsilon, 0.02*\epsilon, 0.05*\epsilon$, respectively.}
  \label{fig:8}
\end{figure*}

In the experiments (\figurename~\ref{fig:8}), we mainly make comparisons between the Pre-Ncut model and the Pre-NCASTV model. Here the size of image is 160*160, and parameters in the Pre-NCASTV model is: $\lambda = 1$, $\epsilon = 0.001*\lambda$, and $\eta=0.03*\epsilon, 0.005*\epsilon, 0.02*\epsilon, 0.05*\epsilon$, respectively. We can find out from these experiments that our proposed edge-based model have better performance. 

\section{Conclusion}
\label{sec_6}
In this paper, we proposed a generalized nonlinear Ncut-based segmentation model with adaptive similarity and spatial regularization. In our model, the similarity function which comes from EM process can be adaptively updated by the model itself. This way, the results of segmentation could be greatly improved. Moreover, we integrated the regularization technique into Ncut method in a variational framework, which enforces the segmentation boundaries to be spatially smooth and guarantees the robustness of the algorithm under noise. In addition, the regularization can equip the similarity function with some spatial location information, which is beneficial for nature image segmentation.

Though the proposed methods have good performance, it can be further improved. For example, the CPU time of the algorithm is much longer than the traditional Ncut models. This leaves space to design more efficient algorithms.
Besides, our model can be extended to deep learning based segmentation by unrolling the proposed algorithm.
\appendix

\section{Proof of Proposition \ref{pro_dual}}\label{appendix_B0}

{
Proof: By the definition of Fenchel-Legendre transformation of $\mathcal{J}$
\begin{equation}\label{f_dual}
\mathcal{J}^{*}(w)=\max\limits_{u}\{<u,w>-\mathcal{J}(u)\}
\end{equation}
and first-order optimal condition, we have
\begin{equation}\label{ww}
w(x,y)=\frac{e^{u^{*}(x,y)}}{\int_{\Omega}e^{u^{*}(x,y)}dy},
\end{equation}
where $u^{*}$ is the maximizer.}

{
By integrating both sides of (\ref{ww}), we have
$\int_{\Omega}w(x,y)dy=1.$
Define $\mathbb{C}_{1}=\{w:\Omega\times\Omega\rightarrow\mathbb{R}|0\leqslant w(x,y)\leqslant 1,\int_{\Omega}w(x,y)dy=1,\forall x\in\Omega\}$, then
\begin{itemize}
\item When $w\in\mathbb{C}_{1}$. If $w(x,y)>0$, then we can set
$w(x,y)=e^{u^{*}(x,y)}$, and obtain the maximizer $u^{*}(x,y)=\ln w(x,y)$. Substitute it into (\ref{f_dual}), we have
\begin{equation}\label{Jstar}
\mathcal{J}^{*}(w)=\displaystyle\int_{\Omega\times\Omega}w(x,y)\ln w(x,y)dxdy.
\end{equation}
Otherwise, there are some $w(x,y)$ to be $0$, one can check that this expression of $\mathcal{J}^*$ in (\ref{Jstar})  is still correct by interpreting  $0 \ln 0$ as $0$.
\item When $w\notin\mathbb{C}_{1}$. Then 
\begin{itemize}
\item if $\exists w(x_0,y_0)<0$, then let $u(x_0,y_0)=-t$ and $u(x,y)=0$ when $(x,y)\neq(x_0,y_0)$, then
\begin{equation}\label{B_1}
<u,w>-\mathcal{J}(u)= -w(x_0,y_0)t-(|\Omega|-1)\ln|\Omega|-\ln[(|\Omega|-1)+e^{-t}].
\end{equation}
Then (\ref{B_1}) converges to $+\infty$ as $t\rightarrow +\infty$.
\item if $w(x,y)>0$, but $\int_{\Omega}w(x,y)dy\neq 1$. Set $u(x,y)=t$, then
\begin{equation}\label{B_2}
<u,w>-\mathcal{J}(u)=\displaystyle\int_{\Omega} t(\int_{\Omega}w(x,y)dy-1)dx-\int_{\Omega}\ln |\Omega|dx.
\end{equation}
If $\int_{\Omega}w(x,y)dy>1$, (\ref{B_2}) converges to $+\infty$ as $t\rightarrow +\infty$; if $\int_{\Omega}w(x,y)dy<1$, (\ref{B_2}) converges to $+\infty$ as $t\rightarrow -\infty$.
\end{itemize}
That is, if $w\notin \mathbb{C}_{1}$, $\mathcal{J}^{*}(w)=+\infty$.
\end{itemize}
Next, we calculate the Fenchel-Legendre transformation of $\mathcal{J}^{*}(w)$, by definition
\begin{equation}\label{double_dual}
\mathcal{J}^{**}(u)=\max\limits_{w\in\mathbb{C}_{1}}\{\int_{\Omega\times\Omega}u(x,y)w(x,y)dxdy-\int_{\Omega\times\Omega}w(x,y)\ln w(x,y)dxdy\}.
\end{equation}
To prove the convexity of $\mathcal{J}$, one direct method is to calculate the second variational of  $\mathcal{J}$ and show it is semi-positive.
We leave this proof method to readers. Here, we give another proof method, which is equivalent to verify $\mathcal{J}^{**}=\mathcal{J}.$
}

{
Here we adopt Lagrangian multiplier method to optimize (\ref{double_dual}), the related Lagrangian functional can be written as
$$
\mathcal{L}(w,v)=\int_{\Omega\times\Omega}u(x,y)w(x,y)dxdy-\int_{\Omega\times\Omega}w(x,y)\ln w(x,y)dxdy+\int_{\Omega}v(x)(\int_{\Omega}w(x,y)dy-1)dx,
$$
where $v$ is the Lagrangian multiplier function. According to the first-order optimal condition of $\mathcal{L}$ with respect to $w$, we have
\begin{equation*}
w^*(x,y)=e^{u(x,y)}e^{v^*(x)-1},
\end{equation*}
where $(w^*, v^*)$ is the saddle of $\mathcal{L}$.
Since $w^*\in\mathbb{C}_{1}$, that is $\int_{\Omega}w^*(x,y)dy=1$, then we have
\begin{equation*}\label{dual_w}
w^*(x,y)=\frac{e^{u(x,y)}}{\int_{\Omega}e^{u(x,y)}dy},
\end{equation*}
by solving $v^*$.
Substituting $w^*$ into (\ref{double_dual}), we have $\mathcal{J}^{**}=\mathcal{J}$, which means that $\mathcal{J}$ is convex with respect to $u$. The Proposition \ref{pro_dual} is proved completely.
}

\section{Proof of Theorem \ref{thm_1}}\label{appendix_B}

Proof:
\begin{equation*}
\renewcommand\arraystretch{2}{
\begin{array}{l}
\mathcal{J}_1(w,h)=\displaystyle\biggl.\int_{\Omega\times\Omega}\left\{\frac{(I(x)-I(y))^2}{2h^2}+\ln(\sqrt{2\pi}h|\Omega|)\right\}w(x,y)dxdy,\\
\mathcal{J}_2(w)~~~=\displaystyle\biggl.\int_{\Omega\times\Omega}w(x,y)\ln w(x,y)dxdy,\\
\mathcal{J}_3(f,w)=\lambda\displaystyle\biggl.\int_{\Omega\times\Omega}[(k_{\epsilon}*f)(x)-(k_\epsilon*f)(y)]^2w(x,y)dxdy,\\
\mathcal{J}_4(f)~~~=\eta\displaystyle\biggl.\int_{\Omega}||\nabla f(x)||dx.
\end{array}}
\end{equation*}

Obviously, $\mathcal{J}_3(f,w)\geq0,~\mathcal{J}_4(f)\geq0$, $\mathcal{J}_1(w,h)\geq \ln(\sqrt{2\pi}|\Omega|h_{min})|\Omega|^2$. For any $t\geq 0$, $t\ln t\geq-\displaystyle\frac{1}{e}$, combining the constraint of $w$, we can get $\mathcal{J}_2(w)\geq-\displaystyle\frac{|\Omega|^2}{e}$. Therefore, $\mathcal{E}(f,w,h)$ has a lower bound and $\displaystyle\inf_{(f,w,h)\in \mathbb{X}} \mathcal{E}(f,w,h)$ exists.

Denote $\{(f_n,w_n,h_n)\}$ as a minimizing sequence of problem (\ref{pf_ncastv}), then
\begin{equation*}\label{4}
  \mathcal{E}(f_n,w_n,h_n)\rightarrow\inf_{(f,w,h)\in \mathbb{X}}\mathcal{E}(f,w,h).
\end{equation*}

Since $\|w_n\|_{L^\infty(\Omega\times\Omega)}\leq 1 $ and $w_n\in L^{\infty}(\Omega\times\Omega)$, $L^\infty(\Omega\times\Omega)$ is the conjugate of $L^1(\Omega\times\Omega)$ which is separable linear normed space, by the Banach-Alaoglu Theorem, there is a weak-$^*$ convergent subsequence (also denoted as $w_{n}$) and a weak-$^*$ limit $w\in L^\infty(\Omega\times\Omega)$ such that
\begin{equation*}
  w_n\rightharpoonup^{*} w~~~in~~L^\infty(\Omega\times\Omega),
\end{equation*}
that is,~for any $\varphi\in L^1(\Omega\times\Omega)$,
\begin{equation*}
 \renewcommand\arraystretch{2}{
\begin{array}{lll}
  \lim\limits_{n\rightarrow+\infty}\displaystyle\biggl.\int_{\Omega\times\Omega}w_n(x,y)\varphi(x,y)dxdy= \displaystyle\biggl.\int_{\Omega\times\Omega}w(x,y)\varphi(x,y)dxdy.\\
\end{array}}
\end{equation*}

Since $w_n(x,y)\varphi(x,y)\leq \varphi(x,y)~a.e.$ for any $\varphi>0$,
$~\varphi\in L^1(\Omega\times\Omega)$.~By Lebesgue dominated convergence theorem,~we can get
\begin{equation*}
 \renewcommand\arraystretch{2}{
\begin{array}{ccc}
\displaystyle\biggl.\int_{\Omega\times\Omega}\lim\limits_{n\rightarrow+\infty}w_n(x,y)\varphi(x,y)dxdy=\displaystyle\biggl.\int_{\Omega\times\Omega}w(x,y)\varphi(x,y)dxdy,\\
\displaystyle\biggl.\int_{\Omega\times\Omega}\lim\limits_{n\rightarrow+\infty}w_n(y,x)\varphi(x,y)dxdy=\displaystyle\biggl.\int_{\Omega\times\Omega}w(y,x)\varphi(x,y)dxdy,
\end{array}}
\end{equation*}
so~$\int_{\Omega}\int_{\Omega}[w(x,y)-\lim\limits_{n\rightarrow+\infty}w_n(x,y)]\varphi(x,y)dxdy=0$ and $\int_{\Omega}\int_{\Omega}[w(x,y)-w(y,x)]\varphi(x,y)dxdy=0$ hold for any~$\varphi\in L^1(\Omega\times\Omega)$. Furthermore,
~$w(x,y)=\lim\limits_{n\rightarrow+\infty}w_n(x,y)$ and $w(x,y)=w(y,x)~a.e.x\in\Omega$. ~It's easy to verify that ~$0\leq w(x,y)\leq 1~a.e.x\in\Omega$ and $\int_{\Omega}w(x,y)dy=1~a.e.x\in\Omega$. Especially, choosing $\varphi(x,y)=\frac{(I(x)-I(y))^2}{2}$,~we can get $$\lim\limits_{n\rightarrow+\infty} \int_{\Omega\times\Omega}\{\frac{(I(x)-I(y))^2}{2}\}w_n(x,y)dxdy=\int_{\Omega\times\Omega}\{\frac{(I(x)-I(y))^2}{2}\}w(x,y)dxdy.$$
\par
By the constraint of $h$, $\{h_n\}$ is  bounded in $\mathbb{R}$. So there exists a subsequence (relabeled as $n$) such that
$\lim\limits_{n\rightarrow+\infty}h_n=h.$
Combining the fact that $\frac{1}{h^2}$ is a continuous with respect to $h$, we can get
$\lim\limits_{n\rightarrow+\infty}\frac{1}{h_n^2}=\frac{1}{h^2}.$
Denote $\mathcal{J}_{1}^{(1)}(w,h)=\int_{\Omega}\int_{\Omega}\{\frac{(I(x)-I(y))^2}{2h^2}\}w(x,y)dxdy,$ then
\begin{equation*}
\renewcommand\arraystretch{2}{
\begin{array}{lll}
|\mathcal{J}_1^{(1)}(w_n,h_n)-\mathcal{J}_1^{(1)}(w,h)|=|(\mathcal{J}_1^{(1)}(w_n,h_n)-\mathcal{J}_1^{(1)}(w,h_n))+(\mathcal{J}_1^{(1)}(w,h_n)-\mathcal{J}_1^{(1)}(w,h))|\\
\leq |\displaystyle\biggl.\int_{\Omega\times\Omega}\frac{(I(x)-I(y))^2}{2 h^2_{min}}(w_n(x,y)-w(x,y))dxdy|+\|I\|_{L^\infty(\Omega)}^2|\Omega|^{2}|\frac{1}{h^{2}_n}-\frac{1}{h^2}|.
\end{array}}
\end{equation*}
As $n\rightarrow+\infty$, the right hand side of the above inequality is $0$. Hence $\lim\limits_{n\rightarrow+\infty}\mathcal{J}^{(1)}_1(w_n,h_n)=
\mathcal{J}^{(1)}_1(w,h)$.
Denote $\mathcal{J}_1^{(2)}(w,h)=\ln(\sqrt{2\pi}h|\Omega|)\int_{\Omega}\int_{\Omega}w(x,y)dxdy,$ then $\mathcal{J}_1(w,h)=\mathcal{J}_1^{(1)}(w,h)+\mathcal{J}_1^{(2)}(w,h)$. Since $\ln(\sqrt{2\pi}h|\Omega|)$ is continuous with respect to $h$, then $\lim\limits_{n\rightarrow+\infty}\ln(\sqrt{2\pi}h_n|\Omega|)=\ln(\sqrt{2\pi}h|\Omega|).$ Using the method as analysing $\mathcal{J}_1^{(1)}(w,h)$, then we can get $\lim\limits_{n\rightarrow+\infty}\mathcal{J}_1^{(2)}(w_n,h_n)
=\mathcal{J}_1^{(2)}(w,h)$.
Therefore \begin{equation*}\lim\limits_{n\rightarrow+\infty}\mathcal{J}_1(w_n,h_n)=\mathcal{J}_1(w,h).
\end{equation*}

Since $w\ln w$ is a continuous and convex function, $\mathcal{J}_2(w)$ is weak-$^*$ lower semi-continuous with respect to $w$,~i.e.
\begin{equation*}
 \renewcommand\arraystretch{2}{
\begin{array}{lll}
 \displaystyle\varliminf_{n\rightarrow+\infty}\int_{\Omega\times\Omega}w_n(x,y)\ln w_n(x,y)dxdy \geq \int_{\Omega\times\Omega}w(x,y)\ln w(x,y)dxdy,
  \end{array}}
\end{equation*}
which indicates that $\mathcal{J}_2(w_n)\geq \mathcal{J}_2(w).$
\par
Now, we consider the convergence of $\mathcal{J}_3(f_n,w_n)$. Defining $b_n(x,y):=[(k_{\epsilon}*f_n)(x)-(k_\epsilon*f_n)(y)]^2$, it is clear that $\{b_n(x,y)\}$ is uniformly bounded. Next we will consider the uniform boundedness of the sequence $\{\frac{\partial b_n}{\partial x}(x,y)\}$ and  $\{\frac{\partial b_n}{\partial y}(x,y)\}$. Since
$$\frac{\partial b_n}{\partial x}(x,y)=2[(k_\epsilon*f_n)(x)-(k_\epsilon*f_n)(y)][\frac{(\partial k_\epsilon}{\partial x}*f_n)(x)],$$
$$\frac{\partial b_n}{\partial y}(x,y)=-2[(k_\epsilon*f_n)(x)-(k_\epsilon*f_n)(y)][\frac{(\partial k_\epsilon}{\partial y}*f_n)(y)],$$
by Young inequality, we can immediately get that $\{b_n(x,y)\}$ is a bounded sequence in $W^{1,1}(\Omega\times\Omega)$. By Rellich-Kondrachov compactness theorem, there exists a subsequence of $\{b_n(x,y)\}$ (also denoted as $b_n$) and $b\in L^1(\Omega\times\Omega)$ such that $b_n(x,y)\rightarrow b(x,y)$ in $L^1(\Omega\times\Omega)$.\par
$\mathcal{J}_3(f,w)$~is a continuous and convex function with variable $f$, then it is weakly lower semi-continuous,~i.e.
\begin{equation*}
 \renewcommand\arraystretch{2}{
\begin{array}{lll}
   \displaystyle\varliminf\limits_{n\rightarrow+\infty}\int_{\Omega\times\Omega}b_n(x,y)w(x,y)dxdy\geq \displaystyle\int_{\Omega\times\Omega}((k_{\epsilon}
  *f)(x)-(k_\epsilon*f)(y))^2w(x,y)dxdy.
\end{array}}
\end{equation*}
Furthermore,~owing to $w_n\rightharpoonup^*w$ in $L^\infty(\Omega\times\Omega)$,~$b_n\rightarrow b$ in $L^1(\Omega\times\Omega)$,~by some simple calculation,~it is clear to get
\begin{equation*}
   \displaystyle\lim_{n\rightarrow+\infty}\int_{\Omega\times\Omega}b_n(x,y)[w_n(x,y)-w(x,y)]dxdy=0.
\end{equation*}
Note that
\begin{equation*}
 \renewcommand\arraystretch{2}{
\begin{array}{rcl}
  \mathcal{J}_3(f_n,w_n)&=&\lambda \displaystyle\int_{\Omega\times\Omega}b_n(x,y)w_n(x,y)dxdy\\
  &=&\lambda \displaystyle\int_{\Omega\times\Omega}b_n(x,y)[w_n(x,y)- w(x,y)]dxdy+\lambda \displaystyle\int_{\Omega\times\Omega}b_n(x,y)w(x,y)dxdy,
\end{array}}
\end{equation*}
 so we have
\begin{equation*}
\varliminf\limits_{n\rightarrow+\infty}\mathcal{J}_3(f_n,w_n)\geq \mathcal{J}_3(f,w).
\end{equation*}
By the definition of $\{(f_n,w_n,h_n)\}$,~the sequence $\mathcal{}(f_n,w_n,h_n)$ is bounded,~i.e.~there exists a constant $M$ such that $\mathcal{J}_1(w_n,h_n)+\mathcal{J}_2(w_n)+\mathcal{J}_3(f_n,w_n)+\mathcal{J}_4(f_n)\leq M$.~Since $\mathcal{J}_1(w_n,h_n),~\mathcal{J}_2(w_n),~\mathcal{J}_3(f_n,w_n)$ are lower bounded,~we can get $\mathcal{J}(f_n)$ is upper bounded.~Therefore the sequence $\{f_n\}$ is bounded in $BV(\Omega)$ and there exists a subsequence of $\{f_n\}$ (also denoted as $f_n$)~and $f$ in $BV(\Omega)$ such that~$f_n\rightarrow f$ in $BV-weak^*$ ~and ~$f_n\rightarrow f$~in $L^1(\Omega)-$strong,~i.e.
\begin{equation*}
\lim_{n\rightarrow +\infty}\int_{\Omega}f_n(x)dx=\int_{\Omega}f(x)dx,~~
 \varliminf_{n\rightarrow+\infty}\mathcal{J}_4(f_n)\geq \mathcal{J}_4(f).
 \end{equation*}
It's obviously to get$\int_{\Omega}f(x)dx=0.$

We can calculate
\begin{equation*}
 \renewcommand\arraystretch{2}{
\begin{array}{lll}
  \displaystyle|\int_{\Omega}f_n^2(x)dx-\int_{\Omega}f^2(x)dx|\\
 \displaystyle\leq|\int_{\Omega}f_n(x)(f_n(x)-f(x)dx|+|\int_{\Omega}f(x)(f_n(x)-f(x))dx|\\
 \displaystyle\leq \|f_n\|_{L^\infty(\Omega)}|\int_{\Omega}(f_n(x)-f(x))dx|+\|f\|_{L^\infty(\Omega)}|\int_{\Omega}(f(x)-f_n(x))dx|\\
 \displaystyle\leq C|\int_{\Omega}(f_n(x)-f(x))dx|+\|f\|_{L^\infty(\Omega)}|\int_{\Omega}(f(x)-f_n(x))dx|.\\
\end{array}}
 \end{equation*}
 Let $n\rightarrow+\infty$,~we can get
 \begin{equation*}
 \lim\limits_{n\rightarrow+\infty}|\int_{\Omega}f_n^2(x)dx - \int_{\Omega} f^2(x) dx| \leq 0.
 \end{equation*}
Then
 \begin{equation*}
 \int_{\Omega}f^2(x)dx=\lim_{n\rightarrow+\infty}\int_{\Omega}f_n^2(x)dx=1.
 \end{equation*}
 Hence,~$(f,w,h)\in \mathbb{X}$ ~is a solution of NCASTV model, which completes the proof.

\end{document}